\documentclass[preprint,12pt]{elsarticle}

\usepackage{amssymb}
\usepackage{amsmath}

\usepackage{helvet}         
\usepackage{courier}        
\usepackage{type1cm}        
\usepackage{makeidx}   
\usepackage[table,xcdraw, dvipsnames]{xcolor}  
\usepackage{array}
\usepackage{tabularx}
\usepackage{graphicx}        
\usepackage{multicol}        
\usepackage{multirow}
\usepackage[bottom]{footmisc}
\usepackage{pifont}            
\usepackage{pdfpages}
\usepackage[utf8]{inputenc} 
\usepackage[T1]{fontenc}    
\usepackage[english]{babel} 
\usepackage[a4paper, margin=2.5cm]{geometry} 
\usepackage{subcaption}
\usepackage{hyperref}
\usepackage{microtype} 
\usepackage{placeins}  
\usepackage{float}     
\usepackage{tikz}
\usetikzlibrary{
    calc,
    arrows.meta,
    intersections,
    patterns,
    positioning,
    shapes.misc,
    fadings,
    through,
    decorations.pathreplacing
}
\definecolor{ColorOne}{named}{MidnightBlue}
\definecolor{ColorTwo}{named}{Dandelion}
\definecolor{ColorThree}{named}{Plum}
\usepackage{booktabs}

\journal{Medical Image Analysis}

\begin{document}

\begin{frontmatter}




\title{\textbf{Federated Learning for Surgical Vision in Appendicitis Classification:} Results of the FedSurg EndoVis 2024 Challenge}


\author[1]{Max Kirchner\corref{cor1}}
\ead{max.kirchner@nct-dresden.de}
\cortext[cor1]{Corresponding author}
\author[1]{Hanna Hoffmann}
\author[1]{Alexander C. Jenke}
\author[4,6]{Oliver L. Saldanha}
\author[6]{Kevin Pfeiffer}
\author[19]{Weam Kanjo}
\author[7,8]{Julia Alekseenko}
\author[1]{Claas de Boer}
\author[9]{Santhi Raj Kolamuri}
\author[1]{Lorenzo Mazza}
\author[7,8]{Nicolas Padoy}
\author[12]{Sophia Bano}
\author[13,14]{Annika Reinke}
\author[13,14,15,16,17,18]{Lena Maier-Hein}
\author[11,12]{Danail Stoyanov}
\author[3,4,6]{Jakob N. Kather}
\author[5,10]{Fiona R. Kolbinger}
\author[1,2]{Sebastian Bodenstedt}
\author[1,2]{Stefanie Speidel}
\affiliation[1]{organization={Department of Translational Surgical Oncology, National Center for Tumor Diseases (NCT), NCT/UCC Dresden, a partnership between DKFZ, Faculty of Medicine and University Hospital Carl Gustav Carus, TUD Dresden University of Technology, and Helmholtz-Zentrum Dresden-Rossendorf (HZDR)}, city={Dresden}, country={Germany}}
\affiliation[2]{organization={Centre for Tactile Internet with Human-in-the-Loop (CeTI), TUD Dresden University of Technology}, city={Dresden}, country={Germany}}
\affiliation[3]{organization={Department of Medicine I, Faculty of Medicine and University Hospital Carl Gustav Carus, TUD Dresden University of Technology}, city={Dresden}, country={Germany}}
\affiliation[4]{organization={Medical Oncology, National Center for Tumor Diseases (NCT), University Hospital Heidelberg}, city={Heidelberg}, country={Germany}}
\affiliation[5]{organization={Weldon School of Biomedical Engineering, Purdue University}, city={West Lafayette}, state={IN}, country={USA}}
\affiliation[6]{organization={Else Kroener Fresenius Center for Digital Health, Faculty of Medicine and University Hospital Carl Gustav Carus, TUD Dresden University of Technology}, city={Dresden}, country={Germany}}
\affiliation[7]{organization={IHU Strasbourg, Institute of Image-Guided Surgery}, city={Strasbourg}, country={France}}
\affiliation[8]{organization={University of Strasbourg, CNRS, INSERM, ICube, UMR7357}, city={Strasbourg}, country={France}}
\affiliation[9]{organization={Dr. NTR University of Health Sciences}, city={Vijayawada}, country={India}}
\affiliation[10]{organization={Department of Visceral, Thoracic and Vascular Surgery, Faculty of Medicine and University Hospital Carl Gustav Carus, TUD Dresden University of Technology}, city={Dresden}, country={Germany}}
\affiliation[11]{organization={Digital Technologies, Medtronic}, city={London}, country={United Kingdom}}
\affiliation[12]{organization={UCL Hawkes Institute and Department of Computer Science, University College London}, city={London}, country={UK}}
\affiliation[13]{organization={German Cancer Research Center (DKFZ) Heidelberg, Division of Intelligent Medical Systems}, city={Heidelberg}, country={Germany}}
\affiliation[14]{organization={German Cancer Research Center (DKFZ) Heidelberg, Helmholtz Imaging}, city={Heidelberg}, country={Germany}}
\affiliation[15]{organization={National Center for Tumor Diseases (NCT) Heidelberg, a partnership between DKFZ and Heidelberg University Hospital}, city={Heidelberg}, country={Germany}}
\affiliation[16]{organization={Faculty of Mathematics and Computer Science, Heidelberg University}, city={Heidelberg}, country={Germany}}
\affiliation[17]{organization={Helmholtz Information and Data Science School for Health}, city={Heidelberg, Karlsruhe}, country={Germany}}
\affiliation[18]{organization={Heidelberg University Hospital, Surgical Clinic, Surgical AI Research Group}, city={Heidelberg}, country={Germany}}
\affiliation[19]{organization={Krankenhaus St. Joseph-Stift Dresden GmbH}, city={Dresden}, country={Germany}}

\begin{abstract}
Developing generalizable surgical AI requires multi-institutional data, yet privacy constraints preclude direct data sharing, making Federated Learning (FL) a natural candidate. Its application to complex, spatiotemporal surgical video remains largely unbenchmarked. We present the FedSurg Challenge, the first international initiative dedicated to FL in surgical vision, as a proof-of-concept evaluation using a multi-center dataset of laparoscopic appendectomies (preliminary subset of Appendix300). Three participant submissions were evaluated on generalization to an unseen clinical center and center-specific local adaptation, alongside centralized, Swarm Learning and parameter-efficient fine-tuning baselines, and constant and chance-level reference classifiers. Our analysis identifies temporal modeling as the architectural factor most consistently associated with generalization to the unseen center, although the effect is not uniform across metrics in the controlled baseline comparison, and both parameter-efficient baselines remain more costly than a constant predictor under the ordinal metric. Classifier collapse arises both from failure of the global model to transfer under domain shift and from unconstrained fine-tuning on small, imbalanced local datasets, motivating structured personalized FL with parameter-efficient fine-tuning for center-specific adaptation. Absolute performance remains far from clinical viability: even with all data pooled centrally the task reached 26.31\% F1-score on the unseen center, and no clinically acceptable threshold for intraoperative appendicitis grading has been established. Paired permutation tests over the test cases resolve only large differences, and no adaptation comparison reaches significance at this sample size. By characterizing these limitations, and the limits of what the evaluation can resolve at this scale, this work establishes a methodological reference point for privacy-preserving surgical video AI.
\end{abstract}

\begin{graphicalabstract}
\includegraphics[width=\textwidth]{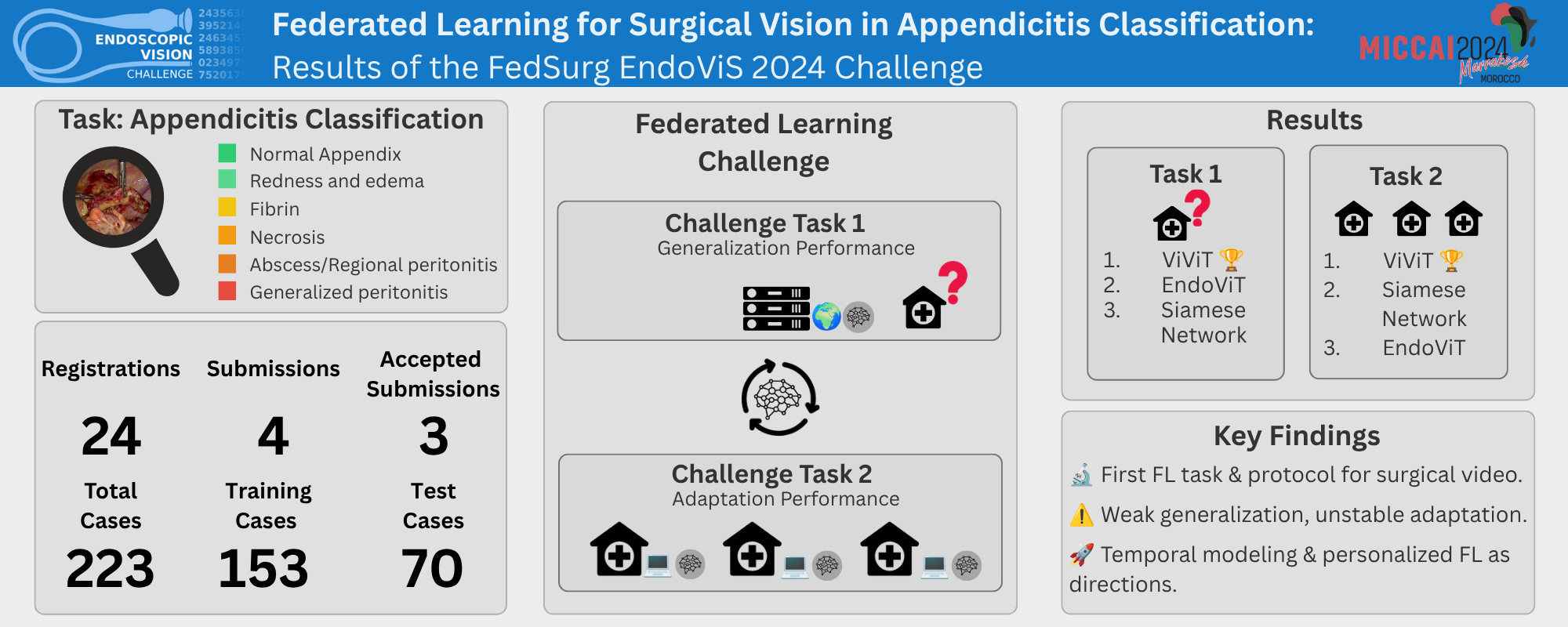}
\end{graphicalabstract}

\begin{highlights}
\item First federated learning challenge in surgical AI  
\item First use of preliminary multi-center Appendix300 dataset  
\item Novel patient-level video classification beyond image-based tasks  
\item Analysis of the generalization vs. personalization trade-off in FL 
\item Temporal modeling and personalized FL as methodological directions
\end{highlights}

\begin{keyword}
Federated Learning \sep EndoVis Challenge \sep Appendectomy \sep Video Classification \sep Surgical Data Science


\end{keyword}

\end{frontmatter}



\section{Introduction}
\label{sec:1}
The combination of early successes in AI algorithms and the emerging field of Surgical Data Science (SDS) holds strong potential to transform surgery \cite{maier-hein_surgical_2017}. Recent work has demonstrated that AI can reliably analyze surgical video, comprehending anatomy, tool usage, and procedural events in real time \cite{brandenburg_active_2023}. Such capabilities are clinically relevant given the established relationship between video-derived quality indicators and postoperative complications. However, the development of robust AI systems critically depends on access to large volumes of high-quality, diverse surgical data. This requirement remains one of the field's most pressing challenges \cite{maier-hein_surgical_2022}.

Surgical datasets are typically sourced from either a single institution or a small group of collaborators \cite{maier-hein_surgical_2022, carstens_artificial_2025}. While these datasets enable preliminary developments, they inherently suffer from limited diversity and scale if applied in real-world scenarios \cite{kirtac_surgical_2022}. Single-center datasets often lack the heterogeneity required for generalizable AI models and involve prolonged data acquisition cycles \cite{lavanchy_challenges_2023}. Multi-institutional data aggregation, on the other hand, can significantly enrich the dataset, but is frequently obstructed by stringent regulatory frameworks such as the Health Insurance Portability and Accountability Act (HIPAA) \cite{rights_ocr_health_2021} or the General Data Protection Regulation (GDPR) \cite{noauthor_general_nodate} that prohibit unrestricted data sharing.

A viable solution is Federated Learning (FL). FL enables decentralized model training across multiple clients, such as hospitals, without the need to share raw image or video data \cite{mcmahan_communication-efficient_2017}. Instead, the model is brought to the data, preserving privacy and complying with regulations like GDPR or HIPAA.
In a standard FL workflow, each client trains a model locally on its private dataset. Only model updates are sent to a central server, where they are aggregated using strategies like Federated Averaging (FedAvg) \cite{mcmahan_communication-efficient_2017} or more robust alternatives such as FedMedian \cite{yin_byzantine-robust_2018}. The updated global model is then redistributed to clients, and this cycle repeats until convergence.

FL is especially well-suited for medical applications, where data privacy, security, and the lack of standardized datasets pose significant challenges \cite{rieke_future_2020}. It enables large-scale collaborative model development without the need to centralize sensitive, multi-institutional data, thereby preserving patient confidentiality and regulatory compliance \cite{rieke_future_2020}. However, FL also introduces new challenges. These include handling data and system heterogeneity across institutions \cite{li_federated_2020, kairouz_advances_2021, rauniyar_federated_2023}, balancing personalization with generalization \cite{tan_towards_2023}, managing communication overhead due to frequent model updates \cite{mcmahan_communication-efficient_2017, kairouz_advances_2021}, addressing fairness and potential model biases \cite{kairouz_advances_2021, li_fair_2020}, and coping with the increased complexity of evaluation and debugging in distributed settings \cite{kairouz_advances_2021}. In the context of SDS, recent studies have begun to explore FL for tasks such as surgical phase recognition, scene segmentation, and tool detection \cite{kassem_federated_2022, kirchner_federated_2025, li_ultraflwr_2025}.

To address challenges in SDS, initiatives such as the \href{https://endovis.org}{Endoscopic Vision (EndoVis) Challenge} \cite{speidel_endoscopic_nodate} have become critical accelerators for progress. The Federated Learning for Surgical Vision (FedSurg) challenge, introduced as part of the 2024 edition of EndoVis, represents the first FL challenge in the field of SDS. The challenge aimed to establish a new classification task, dataset and evaluation protocol for applying FL to surgical AI, enabling comparison of privacy-preserving model development strategies across institutions. We use the term benchmark in the sense of a shared task definition with a fixed dataset split and a specified evaluation and ranking protocol, rather than in the sense of a mature comparison among methods already close to state-of-the-art performance. With three complete submissions, this first edition establishes the former and does not claim the latter; the evaluation accordingly characterizes the limits of what the protocol can currently resolve as well as the relative performance of the submitted methods.

In particular, the challenge focused on evaluating model performance in terms of generalization to unseen centers versus adaptation to individual centers, reflecting two core challenges of FL in real-world surgical applications. 
The Appendix300 dataset is a multi-institutional appendectomy video dataset \cite{kolbinger_appendix300_2025}. As the first FL challenge of its kind in SDS, FedSurg pioneers the use of a preliminary version of the Appendix300 dataset as a foundational platform for algorithm development and validation. A separate study using the complete dataset is published \cite{benchmarking_paper}. In this paper, we report the challenge design, results, and findings of FedSurg according to the transparent reporting of biomedical image analysis challenges (BIAS) guidelines \cite{maier-hein_bias_2020}.

The main contributions of this paper are:

\begin{itemize}
    \item The design and results of FedSurg, the first international challenge in Surgical Data Science dedicated to Federated Learning, which pioneers the use of the preliminary, multi-center Appendix300 dataset.

    \item The first shared task definition and evaluation protocol for FL on patient-level surgical video classification, with the three submitted methods and four organizer baselines evaluated against constant and chance-level references.

    \item A rigorous, data-driven analysis of the critical trade-off between model generalization to unseen institutions and personalization to local client data.
    
    \item A characterisation of the practical limitations and failure modes of FL in surgical AI, contextualised against constant and chance-level reference classifiers and assessed by bootstrap rank stability and paired permutation testing.

\end{itemize}

\section{Challenge Design}
This section outlines the design of the FedSurg challenge, detailing its organizational structure, the core mission guiding the competition, and the datasets employed for evaluation. Additionally, it describes the assessment methods used to fairly and rigorously evaluate submitted models and FL settings, ensuring robust comparison across the multi-centric surgical video dataset.

\subsection{Challenge Organization}
The FedSurg challenge was a one-time event with a fixed submission deadline, held as part of the Medical Image Computing and Computer Assisted Intervention (MICCAI) 2024 conference in Marrakech, Morocco (October 6–10, 2024). Organized jointly by research groups from the Dresden University of Technology (TUD), Purdue University, and the National Center for Tumor Diseases (NCT) Dresden (see~\ref{A:challenge_orga}), the challenge offered a prize pool of €500, provided by the Horizon Europe NearData project and distributed equally between the two challenge tasks.

The primary mission of the FedSurg challenge was to benchmark FL approaches for surgical video classification using a new multi-institutional dataset. The dataset was collected from four German hospitals under institutional ethical approval (see~\ref{A:ethics}). Only training data was released to participants, while test data remained private and was accessible only to the organizers. The dataset used in the FedSurg challenge is a preliminary subset of the Appendix300 dataset \cite{kolbinger_appendix300_2025} (see Subsection \ref{sec:data}). 

\begin{figure}[htbp]
    \centering
    \includegraphics[width=\linewidth]{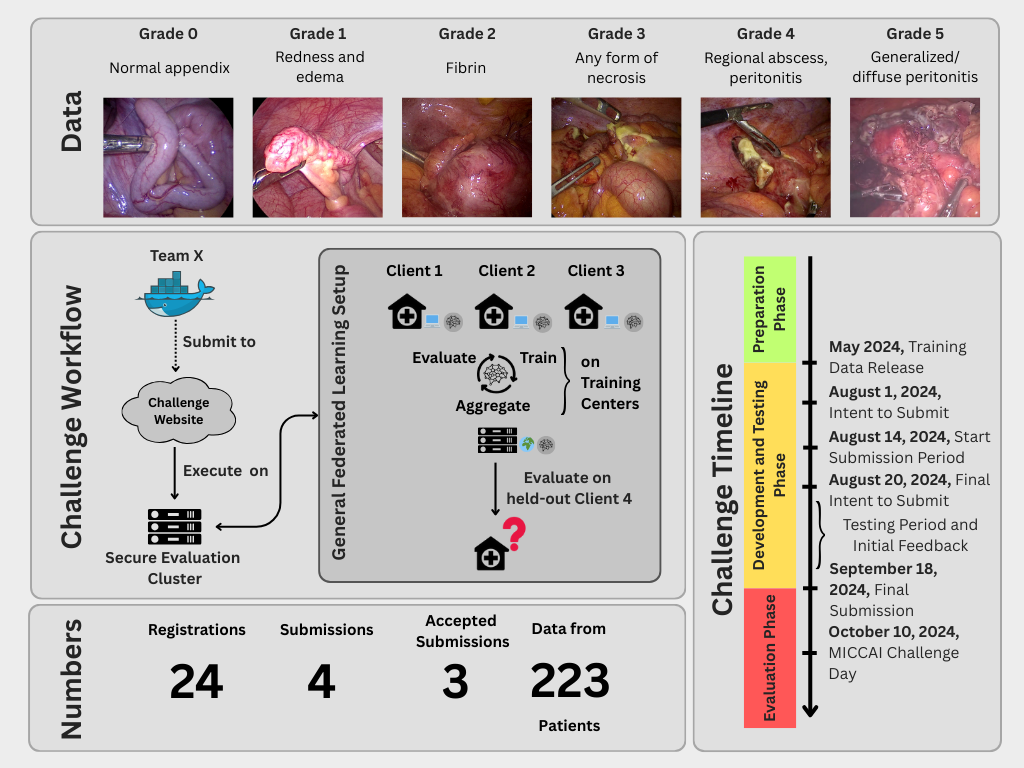}
    \caption{\textbf{FedSurg24 Challenge Highlights:} The top panel shows example images of intraoperative appendicitis grades, defined according to Gomes et al. \cite{gomes_laparoscopy_2012}, which were used for video annotation. The lower panel illustrates the FedSurg Challenge workflow: teams submitted Docker containers via Synapse, which were executed on a secure cluster simulating FL across three centers with local training and centralized aggregation. Final performance was assessed by testing each center’s best local model on its own test set, while the global model was evaluated on the unseen hold-out center to measure generalization. The challenge timeline with key dates is shown alongside.}
    \label{fig:challenge_overview}
\end{figure}

All relevant information, including registration, data access, submission progress tracking, guidelines, and a discussion forum, was made available through the official challenge website on Synapse (see~\ref{A:submission_process}). This platform served as the central hub for participant onboarding, communication, and submission management. Participants were required to sign challenge rules before they could participate and get access to the data (see~\ref{A:rules}). The challenge timeline included registration in April 2024, a feedback-based testing phase during August–September, and the final submission deadline on September 18, 2024 (Figure \ref{fig:challenge_overview}). 

Participants were restricted to the released training data and publicly available resources, including pre-trained models. Challenge-provided data could not be used for pre-training in FL submissions. Members of the organizing institutes were permitted to participate but were not eligible for awards. More information about the challenge is available in the challenge design document (see~\ref{A:design_doc}).

Submissions were required in a containerized format (Docker) and were evaluated on a dedicated server equipped with up to 8 NVIDIA V100 GPUs, 56 CPUs, and 756 GB RAM. This standardized hardware environment and containerized execution ensured reproducibility and fair benchmarking under identical conditions (Figure \ref{fig:challenge_overview}). To support participants, example FL code and evaluation scripts were made publicly available (see \ref{A:submission_process}).

Results were first announced during the MICCAI 2024 Satellite Events and subsequently published on the challenge website, alongside the evaluation framework, rankings, and key performance metrics. All results and analyses from teams with a complete working submission are included in this joint publication, with contributing team members listed as co-authors. Authors were not permitted to publish individual challenge results prior to the release of this paper, ensuring coordinated dissemination and preserving the novelty of the outcomes.

\subsection{Challenge Mission}

The FedSurg challenge focuses on classifying the inflammatory stage of acute appendicitis using laparoscopic appendectomy videos, an important and commonly performed surgical procedure. The primary objective is to provide a comparative benchmark of existing solutions, with a special emphasis on exploring various FL strategies. In particular, the challenge investigates the balance between personalization and generalization under inherent data heterogeneity. For this purpose, the newly created Appendix300 dataset was introduced, with the challenge of utilizing a partial subset of its data. This makes FedSurg not only the first FL challenge in SDS but also the first shared task definition and evaluation protocol for comparing FL approaches on the patient-level task of appendicitis classification.

The challenge cohort comprises human patients, both adults and children of diverse ages and biological sex, who underwent appendectomy for suspected appendicitis at four German medical centers. In general, the challenge cohort is a representative sample of patients undergoing appendectomy for suspected appendicitis. The dataset consists of laparoscopic video recordings captured during these interventions. While no additional patient data was provided for the challenge, the final Appendix300 dataset includes supplementary clinical information \cite{kolbinger_appendix300_2025}.

This challenge's main contribution is its patient-level prediction task for appendicitis staging, a transferable innovation for AI in surgery. This challenge serves as a proof-of-concept for applying FL to patient-level surgical video analysis, with methodological insights intended to transfer to future tasks with broader clinical relevance, such as intraoperative decision support or quality control across diverse surgical procedures.

Participants were tasked with developing FL algorithms to classify appendicitis stages in laparoscopic videos, structured around two core tasks:

\begin{itemize}
    \item Task 1, Generalization: Evaluate the model's ability to generalize to unseen centers. Participants train their models on a subset of centers and are evaluated on a held-out center that was not involved in training.
    \item Task 2, Adaptation: Assess the model’s ability to personalize to each center’s test data. Here, the same trained model is fine-tuned (or adapted) for each single center of the federated setup and then evaluated independently on each center’s test set.
\end{itemize}

In the context of FL, generalization is the ability of a collaborative global model to perform well on data from new clients, while personalization involves adapting the model to achieve optimal performance for a specific client's unique data. The challenge promotes the creation of privacy-preserving algorithms that can tackle both of these issues, creating models that are both robust and adaptable to diverse surgical settings.

\subsection{Challenge Dataset}\label{sec:data}

The challenge dataset is a preliminary subset of the Appendix300 collection and comprises frames from 223 full-length recordings of laparoscopic appendectomies. Beyond image frames and laparoscopic grading, the finalized Appendix300 dataset is enriched with detailed histopathological findings and patient anamnesis data \cite{kolbinger_appendix300_2025}. For this challenge, up to 200 frames were extracted per video using FFmpeg software \cite{tomar2006converting}, capturing a 100-second window sampled at two frames per second around an annotated timestamp \cite{kolbinger_appendix300_2025}. Where the available footage around the timestamp was shorter than this window, fewer frames were obtained.

Frames were selected at the timestamp identified by the operating surgeon when the appendix was fully visible prior to dissection \cite{kolbinger_appendix300_2025}. Inflammation severity, graded from level 0 to 5, was annotated by the operating surgeon (surgery residents with varying years of experience).
The annotation protocol (available at \cite{kolbinger_appendix300_2025}) defines class descriptions and includes illustrative examples based on the definition by Gomes et al. \cite{gomes_laparoscopy_2012}. Furthermore, fine-grained classes 3A/3B were merged into class 3, and classes 4A/4B into class 4 in this challenge dataset. An example overview of the data is visible in Figure \ref{fig:challenge_overview}. By using this annotation protocol, a verbal explanation to the participating surgeons, and a custom graphical user interface of the annotation software, we minimize misclassification. Nonetheless, the reliability of this grading scheme has since been characterised on the complete Appendix300 dataset, of which the challenge data forms a subset. An independent second reading of all recordings yielded a weighted Cohen's kappa of $0.614$ against the original annotation, with exact agreement varying from $0.618$ at grade 0 to no exact agreement at the most severe (grade 5), and nine percent of cases required a third annotation to resolve disagreements of three or more grades \cite{kolbinger_appendix300_2025}. These figures postdate the challenge and do not describe the labels used here, which derive from the original annotation alone, but they indicate that ambiguity in this grading task is not only confined to adjacent grades.

The challenge dataset comprises data collected from multiple hospitals across Germany, including both university and community hospitals with varied surgical profiles. While the challenge simulates an FL setup, it reflects a realistic scenario since the data originates from four distinct, real-world centers that have been anonymized to protect institutional privacy. The contributing institutions include:

\begin{itemize}
\item Asklepios-ASB Klinik Radeberg
\item University Hospital Carl Gustav Carus Dresden, Department of Pediatric Surgery
\item Krankenhaus St. Joseph-Stift, Dresden
\item St. Elisabethen-Krankenhaus, Ravensburg
\end{itemize}

An overview of the video distribution across centers is shown in Table \ref{tab:data_distribution_table}.

\begin{table}[htbp]
\centering
\caption{Data Distribution Across Participating Centers. Video snippets span up to 100 seconds, centered on the timestamp of full appendiceal visibility prior to surgical preparation; where the available footage around the timestamp is shorter, the snippet is correspondingly shorter.}
\label{tab:data_distribution_table} 
\begin{tabular}{cccc}
\hline
\textbf{Center} & \textbf{Training Videos} & \textbf{Testing Videos} & \textbf{Total} \\ \hline
1               & 40                       & 10                      & 50             \\
2               & 33                       & 9                       & 42             \\
3               & 80                       & 22                      & 102            \\
4               & 0                        & 29                      & 29             \\ \hline
Total           & 153                      & 70                      & 223            \\ \hline
\end{tabular}
\end{table}

\begin{figure}[htbp]
    \centering 

    \begin{subfigure}[b]{\textwidth}
        \centering
        \includegraphics[width=0.75\textwidth]{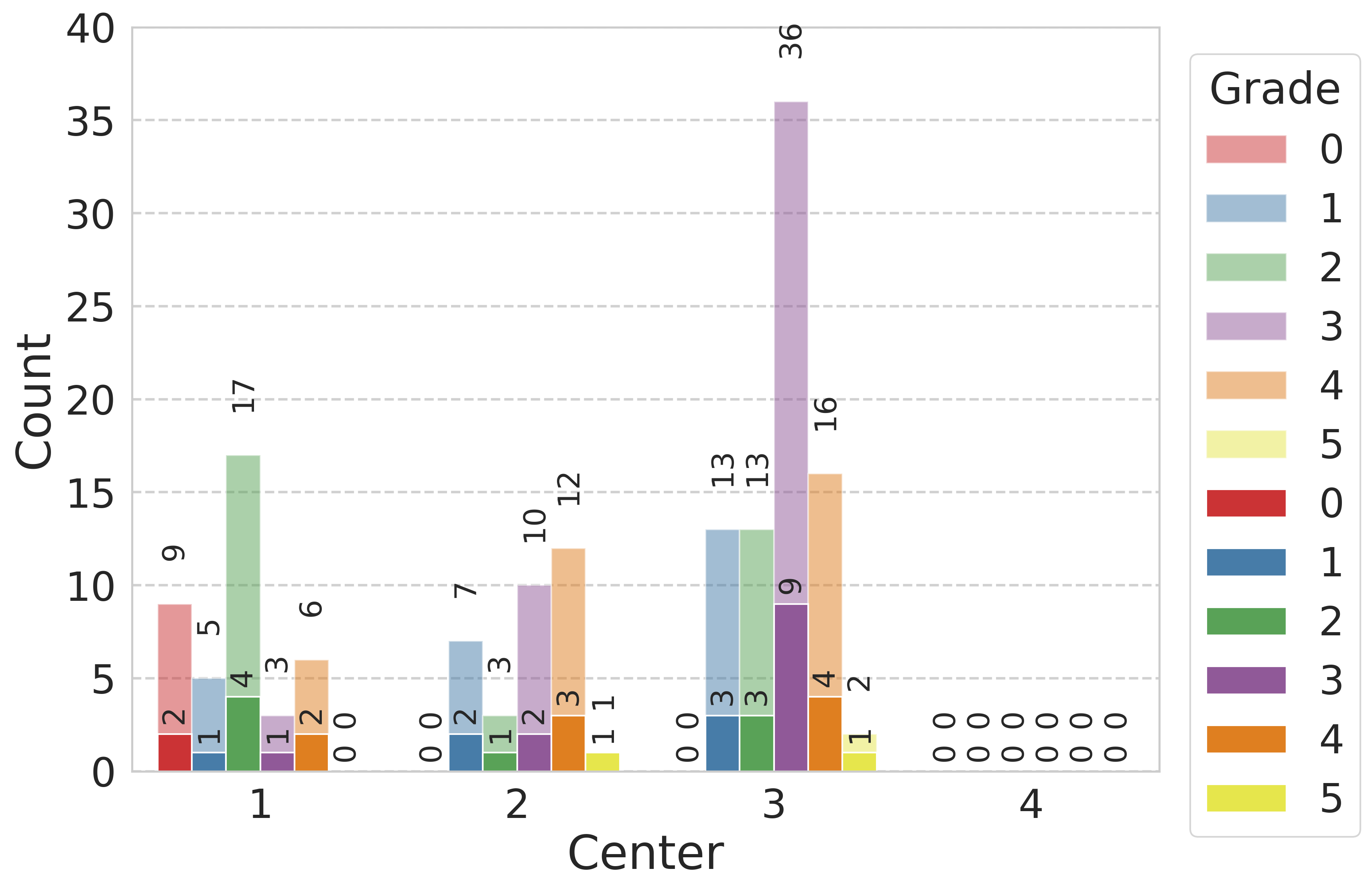}
        \caption{Training Dataset Distribution.}
        \label{fig:sub_a}
    \end{subfigure}

    \begin{subfigure}[b]{\textwidth}
        \centering
        \includegraphics[width=0.75\textwidth]{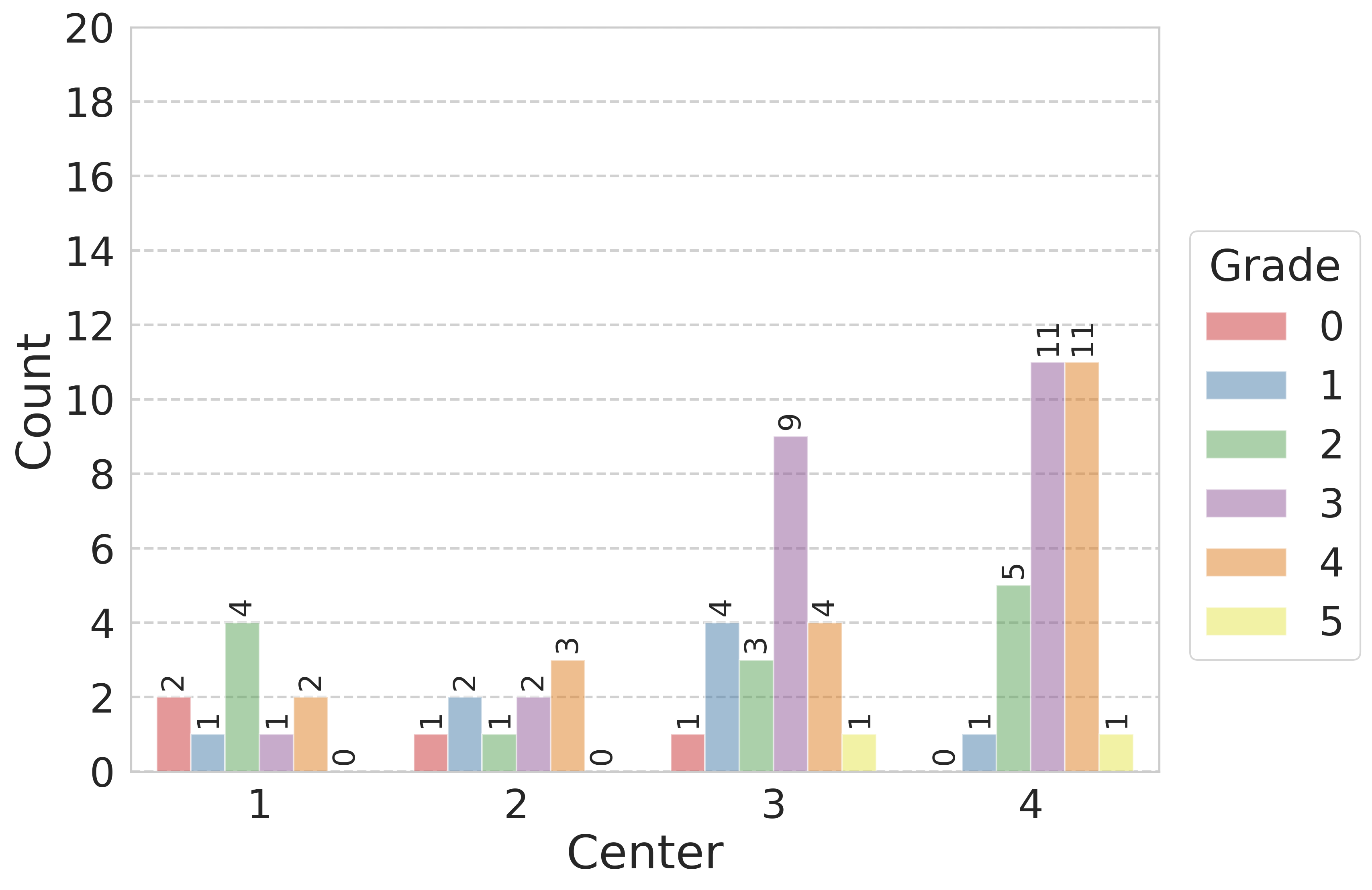}
        \caption{Testing Dataset Distribution.}
        \label{fig:sub_b}
    \end{subfigure}

      \caption{\textbf{Label Distribution Across Data Subsets per Center.} Label distributions for (a) the training dataset and (b) the test dataset across the four centers. In the training set visualization, the darker segments represent the publicly available subset for participant development, while the lighter segments show the complete dataset used for final federated training. The plots highlight notable inter-center variability and class imbalance.}
  \label{fig:data_distribution}
\end{figure}

To discourage centralized model development and preserve the integrity of the federated setting, the training data was partitioned into public and private subsets. Participants were granted access only to the public subset, which constituted 25\% of the total training data. The remaining 75\% was used exclusively by the organizers during the final training phase, where both public and private data were jointly leveraged in a secure federated setup (Figure \ref{fig:challenge_overview} and Figure \ref{fig:data_distribution}). The class distribution within the dataset mirrors real-world clinical prevalence, featuring a predominance of mid-level cases and a lower representation of extreme or mild inflammation levels.

No predefined validation set was provided. Participants were expected to devise their own validation strategies. Final submissions were executed centrally by the organizers, who conducted full federated training and evaluation across all centers under secure and standardized conditions to ensure reproducibility and fairness.

Since the challenge dataset constitutes a preliminary subset of the Appendix300 dataset, we provide a CSV file specifying which samples were used in the challenge and their assignment to centers and to the training and test splits, together with two recordings that were used in the challenge but excluded from the final dataset because the available footage was shorter than the nominal 100 second window or the appendix was not clearly visible at the annotated timestamp. This material is deposited under the Creative Commons Attribution (CC BY) license\footnote{Available at \href{https://doi.org/10.25532/OPARA-1534}{https://doi.org/10.25532/OPARA-1534}}. It does not contain the appendectomy recordings themselves, apart from the two excluded cases; the video data and accompanying clinical metadata are available separately as the Appendix300 dataset, also under CC BY\footnote{Available at \href{https://doi.org/10.25532/OPARA-1173}{https://doi.org/10.25532/OPARA-1173}}. Any use of the challenge dataset requires citing both this paper and the final Appendix300 publication.

\subsection{Assessment Methods}\label{sec:assessment}

The evaluation of this challenge is based on two complementary metrics: the macro-averaged F1-score, also known as the Dice coefficient, and the Expected Cost (EC) with linear weights. Together, these metrics provide a robust assessment of model performance in a multi-class setting with ordinal structure.

\subsubsection*{F1-score (Dice Coefficient):}  
The F1-score is widely used for balancing precision and recall, offering a harmonic mean that reflects both the accurate identification of relevant instances and the minimization of false positives \cite{dice_measures_1945, maier-hein_metrics_2024}. Given a confusion matrix $\mathbf{M} \in \mathbb{N}^{C \times C}$, where $M_{i,j}$ denotes the number of samples with ground-truth class $i$ predicted as class $j$, the F1-score for class $c$ is computed as:

\begin{equation}
\text{F1}_c = \frac{2 \cdot \text{TP}_c}{2 \cdot \text{TP}_c + \text{FP}_c + \text{FN}_c}
\end{equation}

where:
\begin{align*}
\text{TP}_c &= M_{c,c} \quad \text{(true positives)} \\
\text{FP}_c &= \sum_{\substack{i=1 \\ i \neq c}}^{C} M_{i,c} \quad \text{(false positives)} \\
\text{FN}_c &= \sum_{\substack{j=1 \\ j \neq c}}^{C} M_{c,j} \quad \text{(false negatives)}
\end{align*}

The overall F1-score is calculated as the macro-average across all classes:

\begin{equation}
\text{F1}_{\text{macro}} = \frac{1}{C} \sum_{c=1}^{C} \text{F1}_c
\end{equation}

\subsubsection*{Expected Cost (EC)} 
To account for the ordinal nature of the classification task, we also report the Expected Cost (EC), which penalizes misclassifications based on their severity. This aligns with the principle of ordinal monotonicity, where predictions farther from the ground-truth class incur higher penalties \cite{hastie2009elements, elkan2001foundations, maier-hein_metrics_2024}. The EC is defined as:

\begin{equation}
\text{EC} = \frac{1}{N} \sum_{i=1}^{C} \sum_{j=1}^{C} M_{i,j} \cdot w_{i,j}
\end{equation}

where:
\begin{align*}
    N &= \sum_{i=1}^{C} \sum_{j=1}^{C} M_{i,j} && \text{(total number of samples)} \\
    w_{i,j} &= \frac{|i - j|}{C - 1} && \text{(cost of predicting class $j$ when the true class is $i$)}
\end{align*}

with the linear weight function:

\begin{equation}
w_{i,j} = \frac{|i - j|}{C - 1}
\end{equation}

This assigns zero cost to correct predictions and a linearly increasing penalty for deviations, with a maximum cost of 1 for the farthest misclassifications.

\subsubsection{Ranking}\label{sec:ranking}
To ensure a fair evaluation of submissions across both generalization and adaptation scenarios, we implemented a ranking framework that combines multiple performance metrics with an assessment of ranking stability and statistical comparison. Only submissions that successfully completed both tasks were considered. Teams failing to meet this criterion or submitting non-executable code were disqualified. Our ranking methodology guaranteed that models were evaluated not just on average performance but also on their robustness and consistency across heterogeneous data settings. The details of our evaluation procedure are described below.

For Task 1 (generalization), we computed the F1-score and the EC metric for all test cases. Separate rankings were assigned based on each metric, and a team's overall rank for Task 1 was derived by averaging these two ranks. For Task 2 (adaptation), the metrics were first averaged across all three centers, and the same ranking scheme used in Task 1 was then applied.

The final leaderboard was determined by averaging each team's ranks across both tasks, providing a comprehensive assessment of overall performance throughout the challenge.

Additionally, to ensure the robustness and reliability of the rankings, bootstrapping \cite{efron1992bootstrap} was applied. Bootstrapping, as emphasized by Maier-Hein et al. \cite{maier-hein_why_2018}, is a key method for assessing the variability of rankings and the stability of observed performance differences. In this study, the test set was repeatedly resampled with replacement for 10,000 iterations, and team rankings were recalculated for each resample. For each team, we computed the proportion of iterations in which it retained its original rank as well as the proportions in which it achieved each of the other possible ranks. Resampling indices are drawn once per replicate and shared across all methods, and predictions are aligned by case identifier before resampling, so that all methods are compared on identical resampled sets. To characterize the consistency of observed differences under resampling, we applied the Wilcoxon signed-rank test to the bootstrapped metric values obtained over all iterations. Because these values are computed over resampling replicates rather than over the test cases themselves, their effective sample size is set by the number of iterations rather than by the data, and we therefore report them as stability diagnostics rather than as inferential tests.

To contextualize absolute performance, we additionally evaluate four reference classifiers derived from training labels only. TrivialGlobal predicts the pooled training mode on every case, and TrivialLocal predicts each center's own training mode. RandomUniform samples the six grades with equal probability, and RandomPrior samples according to training-set prevalence; both stochastic references are simulated over 10,000 draws. All four are contextual references and are excluded from every rank computation.

As an inferential complement to the bootstrap analysis, we additionally conduct paired permutation tests over the test cases themselves. The exchangeable unit is the individual case: for each of $10{,}000$ permutations, every case is independently assigned to one of the two methods under comparison with probability $0.5$. For Task 1 the metric is recomputed over the held-out center's test cases; for Task 2 it is recomputed per center from the permuted assignment and the three centers are then averaged with equal weight, matching the ranking estimand defined above. All tests are two-sided, comparing the absolute observed difference against the absolute differences of the permutation distribution. With $10{,}000$ permutations the smallest attainable $p$-value is $10^{-4}$; values reported as $0.0001$ should therefore be read as $p \le 10^{-4}$, and a value of $1.0000$ indicates that the observed difference was equalled or exceeded in essentially every permutation.

\section{Results}\label{sec:results}
The FedSurg challenge received 24 registrations, with four final submissions. However, one submission was disqualified due to non-executable code after the final deadline. This left three complete submissions for evaluation. The participating teams were: Santhi R. Kolamuri, who submitted independently as Team Santhi; Lorenzo Mazza and Claas de Boer from the Translational Surgical Oncology group at the National Center for Tumor Diseases, submitting as Team Elbflorenz; and Julia Alekseenko and Nicolas Padoy from the CAMMA research group at IHU Strasbourg, submitting as Team Camma. Furthermore, we provide a centralized baseline, a federated Swarm Learning (SL) baseline, and two parameter-efficient fine-tuning baselines, together with constant and chance-level reference classifiers.

\subsection{Participating Teams and Methods}
The following section details the methodologies submitted by participants in the FedSurg challenge. Each subsection outlines the architectural choices, training strategies, and FL configurations employed. Where relevant, we contextualize each approach within existing literature on foundation models, metric learning, and federated optimization. A summary of each submission is presented in Table \ref{tab:submissions}.

\begin{figure}[htbp]
    \centering
    \includegraphics[width=\linewidth]{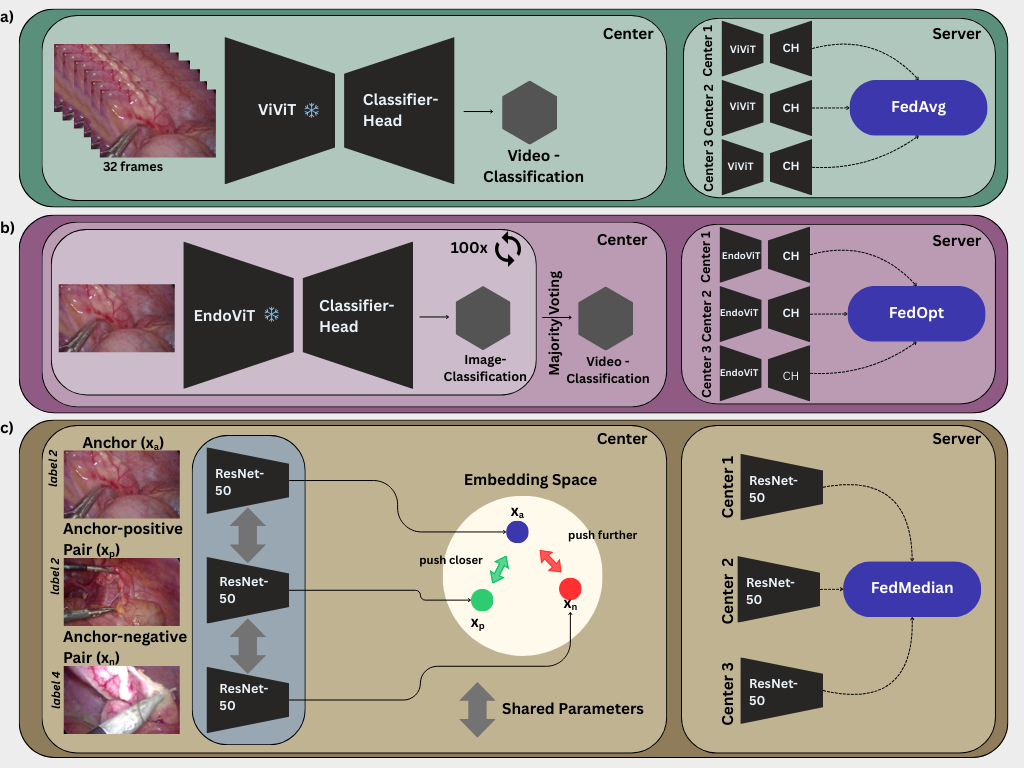}
    \caption{\textbf{Methods Overview:} The three submissions shown utilize different backbone architectures and federated strategies. A common approach is that in each server round, the best-performing model from a client's local training rounds is sent to the server for aggregation. (a) Team Santhi uses a frozen ViViT backbone with a fine-tuned classification head processing 32 frames per video, with updates aggregated via FedAvg. (b) Team Elbflorenz uses a frozen EndoViT backbone with a fine-tuned head, predicting single frames repeatedly and combining them via majority voting, with updates aggregated via FedSAM. (c) Team Camma uses ResNet50 models trained with a contrastive approach on positive and negative pairs, with updates aggregated via FedMedian. At inference, classification is performed by comparing the test embedding to a support set.}
    \label{fig:methods}
\end{figure}

\begin{table}[htbp]
\centering
\small
\caption{Overview of submissions to the FedSurg Challenge, detailing each team's backbone model, prediction method, frame sampling, loss function, optimizer, learning rate, batch size, and federated learning configuration (FL strategy, FL-rounds, and local rounds).}
\label{tab:submissions}
\rowcolors{2}{gray!10}{white}
\begin{tabularx}{\textwidth}{lXXX} 
\hline
\textbf{Team} & \textbf{Santhi} & \textbf{Elbflorenz} & \textbf{Camma} \\
\hline
\textbf{Backbone Model} & Frozen pretrained Video Vision Transformer (ViViT) on Kinetics-400. & Frozen pretrained EndoViT on Endo700k. & Siamese network with a ResNet-50 backbone. \\
\textbf{Prediction Method} & Video-level classification. & Majority voting on independent frame-level classifications. & Compares test embedding to class prototypes or a full support set. \\
\textbf{Frame Sampling} & Samples 32 frames, with two-thirds from a narrow window around the center and the rest from the full sequence with a bias towards the center. & Uses 100 equidistant frames from each video. & Selects representative frames based on cosine similarity to a reference embedding, plus the central keyframe. \\
\textbf{Loss Function} & Cross-entropy loss. & Weighted cross-entropy loss to address class imbalance. & Triplet margin loss with cosine similarity. \\
\textbf{FL Strategy} & Federated Averaging (FedAvg). & Adaptive Federated Sharpness-Aware Minimization (FedSAM) on Client-side and FedOpt on Server Side. & Federated Median (FedMedian). \\
\textbf{Optimizer}         & Adam                                                                                                                                          & adaptive Federated Sharpness Aware Minimization (FedSAM) based on SGD                              & Adam                                                                                                          \\
\textbf{FL-Rounds}         & 20                                                                                                                                             & 50                                                                                                 & 10                                                                                                            \\
\textbf{Local Rounds}      & 5                                                                                                                                            & 2                                                                                                  & 5                                                                                                             \\
\textbf{Learning Rate}     & $1 \times 10^{-4}$                                                                                                                                        &$1 \times 10^{-3}$                                                                                             & $1 \times 10^{-6}$                                                                                                      \\
\textbf{Batchsize}         & 4                                                                                                                                             & 128                                                                                                & 1                                                                                                             \\ 

\hline
\end{tabularx}
\end{table}

\subsubsection{Team Santhi}
Santhi R. Kolamuri’s approach leverages a pretrained Video Vision Transformer (ViViT) \cite{arnab_vivit_2021}, which is well-suited for capturing spatio-temporal features in surgical video sequences (Figure \ref{fig:methods} (a)). ViViT’s transformer-based architecture has proven more effective than conventional CNNs in modeling temporal dependencies. By utilizing ViViT pretrained on Kinetics-400, only the final classification layer is fine-tuned, while all other weights remain frozen. This linear probing approach is common for foundation models \cite{dosovitskiy_image_2021}, as it allows efficient adaptation while retaining robust pretrained representations.

The frame loading strategy is customized to emphasize salient temporal regions. From 200 available video frames, 32 are sampled for training. Two-thirds are selected from a narrow window around frame 100 to capture high-information segments, while the remainder of the frames are drawn from the full sequence with 60\% probability to the frames near the center. This hybrid sampling balances local relevance and temporal diversity.

Training is implemented via the Flower FL framework \cite{beutel_flower_2022}, using FedAvg for aggregation. Centers train locally with a batch size of four for five epochs per round, across 20 communication rounds. Cross-entropy loss is used with mixed precision training to improve memory efficiency. The lightweight design, frozen backbone, focused sampling, and shallow head enable fast convergence with low resource demands. This submission builds on the success of ViViT in centralized video classification while adapting it to the constraints of FL.

\subsubsection{Team Elbflorenz}
Team Elbflorenz adopts a foundation model-based strategy using EndoViT \cite{batic_endovit_2024}, a Vision Transformer pretrained on the Endo700k dataset consisting of diverse endoscopic images (Figure \ref{fig:methods} (b)). Following a standard linear probing setup \cite{dosovitskiy_image_2021}, the encoder is frozen and only a lightweight classification head is trained. This setup is motivated by recent successes of foundation models in medical imaging \cite{yang_large-scale_2025, schmidgall_general_2024}, where pretrained encoders generalize well even with limited task-specific data.

The model uses 100 equidistant frames from each video and independently classifies each frame. A weighted cross-entropy loss is used to mitigate class imbalance, with weights inversely proportional to class frequency. Final video-level predictions are determined via majority voting. In the event of a tie, average confidence scores guide the decision. This per-frame classification strategy increases robustness by aggregating multiple frame-level predictions.

Federated optimization on the client side is handled using adaptive FedSAM \cite{caldarola_improving_2022, foret_sharpness-aware_2021}, which extends Sharpness-Aware Minimization (SAM) to heterogeneous FL settings. This method encourages flatter minima and better generalization across non-IID (independent and identically distributed) client data. Similar to Team Santhi's approach, training was implemented via the Flower FL framework \cite{beutel_flower_2022}. It employs 2 local epochs and 50 global rounds, with center-side class balancing and hyperparameter tuning. Compared to standard FedAvg, FedSAM has demonstrated improved convergence on non-IID data \cite{caldarola_improving_2022}. Model aggregation on the server is done with FedOpt \cite{reddi_adaptive_2021}. By combining EndoViT with adaptive optimization, Team Elbflorenz presents a minimal yet effective pipeline for surgical appendicitis classification under federated constraints.

\subsubsection{Team Camma}
Team Camma employs a metric learning approach based on a Siamese network with a ResNet-50 backbone \cite{he_deep_2015}, inspired by the original Siamese architecture \cite{bromley_signature_1993}. Unlike conventional classifiers that output class probabilities, this model maps input images to L2-normalized 256-dimensional embeddings. Training is performed using triplet margin loss with cosine similarity, encouraging embeddings from the same appendicitis stage to cluster while pushing apart those from different stages (Figure \ref{fig:methods} (c)).

At inference, classification is performed by comparing the test embedding to a support set. This is done either via class prototypes (mean embeddings) or by averaging distances to all embeddings within each class. Camma observed that prototype-based inference worked best for Centers 2 and 3, while per-sample comparison was more effective for Center 1, allowing the approach to flexibly adapt to center-specific data distributions.

This metric learning paradigm is particularly well-suited for FL, as it focuses on learning a robust embedding space rather than simply aggregating classifier weights, which can be highly sensitive to the non-IID label distributions across centers.

To address domain heterogeneity, the model incorporates Switchable Normalization \cite{luo_switchable_2021}, which combines batch, instance, and layer normalization through softmax gating. This adaptive scheme enhances generalization across diverse centers, which is crucial in federated settings.

In addition, frame selection leverages embedding-based similarity: representative frames are chosen based on cosine similarity to a ResNet-50 reference embedding, alongside the keyframe (frame 100), to improve input consistency.

Training is conducted within a custom FL framework using the FedMedian aggregation strategy \cite{yin_byzantine-robust_2018}, which is robust to outliers. Each center performs 5 local epochs of triplet optimization per federation round, across 10 federation rounds (Table~\ref{tab:submissions}). The locally best-performing model (by F1-score) is selected for aggregation, ensuring only high-quality updates contribute to the global model. Overall, Team Camma’s method demonstrates how metric learning, adaptive normalization, and robust aggregation can be combined for scalable, personalized FL in surgical video classification.

\subsection{Baselines}\label{sec:baselines}
In addition to the three participating teams, we provide four baselines
for comparison: a centralized baseline, a federated Swarm Learning baseline, and two
parameter-efficient fine-tuning baselines. The first two are based on the SurgTempoNet
method \cite{benchmarking_paper}.

SurgTempoNet employs a spatial-temporal architecture designed to predict patient-level clinical outcomes directly from laparoscopic videos. The model utilizes a ConvNeXt-Base~\cite{liu2022convnet} backbone, pre-trained on ImageNet-22k~\cite{russakovsky2015imagenet}, to extract high-level feature representations from individual frames. In its temporal configuration, these features are fed into a single-layer Long Short-Term Memory (LSTM)~\cite{hochreiter1997long} network with a hidden state size of $160$. This allows the model to aggregate sequential information across the entire surgical procedure, mapping a sequence of frames to a single global label through a final linear classification layer. 

To manage the high dimensionality of surgical footage, the pipeline uses a weakly supervised approach with sparse frame sampling. Videos are processed at a rate of one frame per second, and frames are resized to a resolution of $480 \times 270$ pixels. This sampling strategy reduces the computational burden of processing every raw video frame. 

Both centralized and federated training is conducted using the AdamW optimizer with a learning rate of $1.0\times 10^{-3}$. To address the inherent class imbalance in surgical datasets, the model utilizes Weighted Cross-Entropy Loss, where weights are inversely proportional to the frequency of each disease stage in the training set. Furthermore, a batch accumulation strategy is implemented, where gradients are computed over mini-batches of $64$ clips, but weights are updated only every three steps, effectively stabilizing the training process for video sequences. 

The defining characteristic of the federated baseline is its integration into a SL framework~\cite{warnat2021swarm, hpe_swarmlearning}. Unlike centralized training or standard FL, local models are trained on site-specific data, and only the model weights are synchronized via a peer-to-peer network managed by a SwarmCallback function. This synchronization occurs at a fixed frequency (e.g., every $100$ batches) and requires a minimum number of active peers to proceed.

Furthermore, to probe whether parameter-efficient fine-tuning (PEFT) constitutes a viable adaptation strategy in this data regime, two additional preliminary models are included. Both apply Low Rank Adaptation (LoRA)~\cite{hu2022lora} (rank~8, alpha~16) to the attention projections of a frozen backbone, combined with a shared three-layer MLP classification head and FedAvg aggregation over 25 rounds; all remaining hyperparameters (ASAM optimizer, cosine annealing, four local epochs per round) are held constant.

The \textit{EndoViT+LoRA} baseline builds on the Team Elbflorenz architecture, adapting the frame-level EndoViT backbone \cite{batic_endovit_2024} (ViT-Base, pretrained on endoscopic image data (Endo700k); $\sim$940K trainable parameters) with single-frame inputs ($224 \times 224$). The \textit{ViViT+LoRA} baseline extends this to temporal modeling by substituting a ViViT-B-16$\times$2 backbone pretrained on Kinetics-400, processing 16-frame clips through native spatial-temporal attention ($\sim$969K trainable parameters), following the architectural principle of Team Santhi. The two backbones therefore differ both in temporal modeling and in pretraining domain, endoscopic for EndoViT and general-purpose video for ViViT. Both baselines are post-hoc organizer analyses and are not part of the official challenge ranking. They are reported as point estimates throughout and are excluded from the bootstrap rank stability analysis, the permutation tests and the tabulated reference-classifier comparisons, all of which cover the three submissions together with the centralized and Swarm Learning baselines; where the two parameter-efficient baselines are compared against a reference classifier in the text, that comparison is made on point estimates alone.

\subsection{Scores and Rankings}
The performance of the submitted federated models was evaluated in both tasks of the FedSurg Challenge: generalization to an unseen center (Task 1) and center-specific adaptation (Task 2). Performance was measured using Expected Cost (EC) and F1-score. Rank stability was assessed via 10,000-iteration bootstrapping, with statistical comparison by paired permutation tests over the test cases.
Additionally, the baseline methods are reported separately in Section \ref{sec:FurtherAnalysis} as contextual references and are not part of the official challenge ranking.
Overall, a comprehensive summary of results, including key metrics, confusion matrices, and rank stability, is provided in Tables \ref{tab:Task1} and \ref{tab:Task2} and Figures \ref{fig:confusion_matrix_1}, \ref{fig:bootstrapping_results}, \ref{fig:rankstability}, and \ref{fig:confusion_matrix_2}, with further details in \ref{A:bootstrapping}. 

\subsubsection{Task 1: Generalization to an Unseen Center}

\begin{table}[htbp]
\centering
\caption{Performance comparison of the three teams on Task 1 at the held-out center (Center 4), reporting Expected Cost (EC, lower is better) and F1-score (higher is better). Best results are highlighted in bold.}
\label{tab:Task1}
\begin{tabular}{lccc}
\hline
Team        &  EC \(\downarrow \)    & F1  \(\uparrow \)    \\ \hline
Camma        &   57.24\% & 4.76\% \\
Elbflorenz   &  24.14\% & 7.83\% \\
Santhi       &  \textbf{12.41\%} & \textbf{23.03\%} \\ \hline
\end{tabular}
\end{table}

\begin{figure}[htbp]
    \centering
    \includegraphics[width=\linewidth]{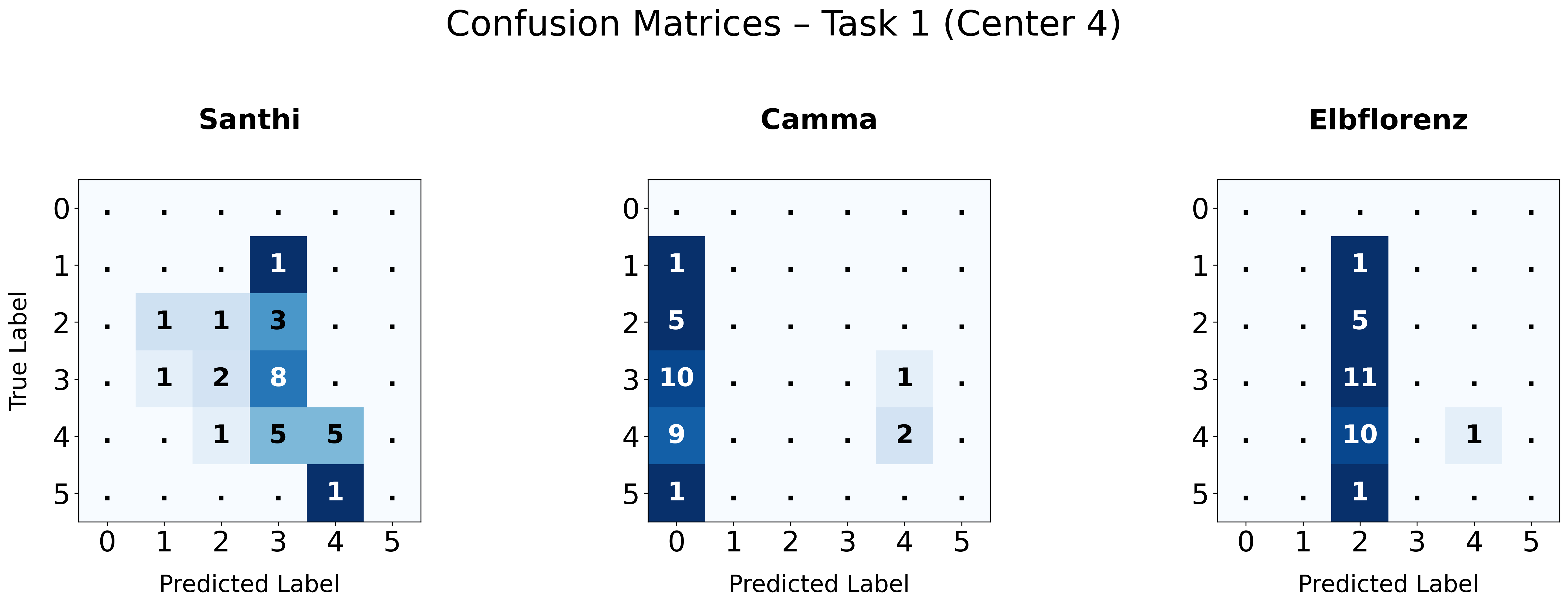}
    \caption{\textbf{Confusion Matrices – Task 1, Center 4.} Confusion matrices for the participating teams on Center 4 (Task 1). The values in the confusion matrices are not normalized. The color highlighting is normalized row-wise by true labels.  The diagonal highlights class-wise recall, while off-diagonal values indicate common misclassification patterns.
}
    \label{fig:confusion_matrix_1}
\end{figure}

\begin{figure}[htbp]
    \centering
    \includegraphics[width=\linewidth]{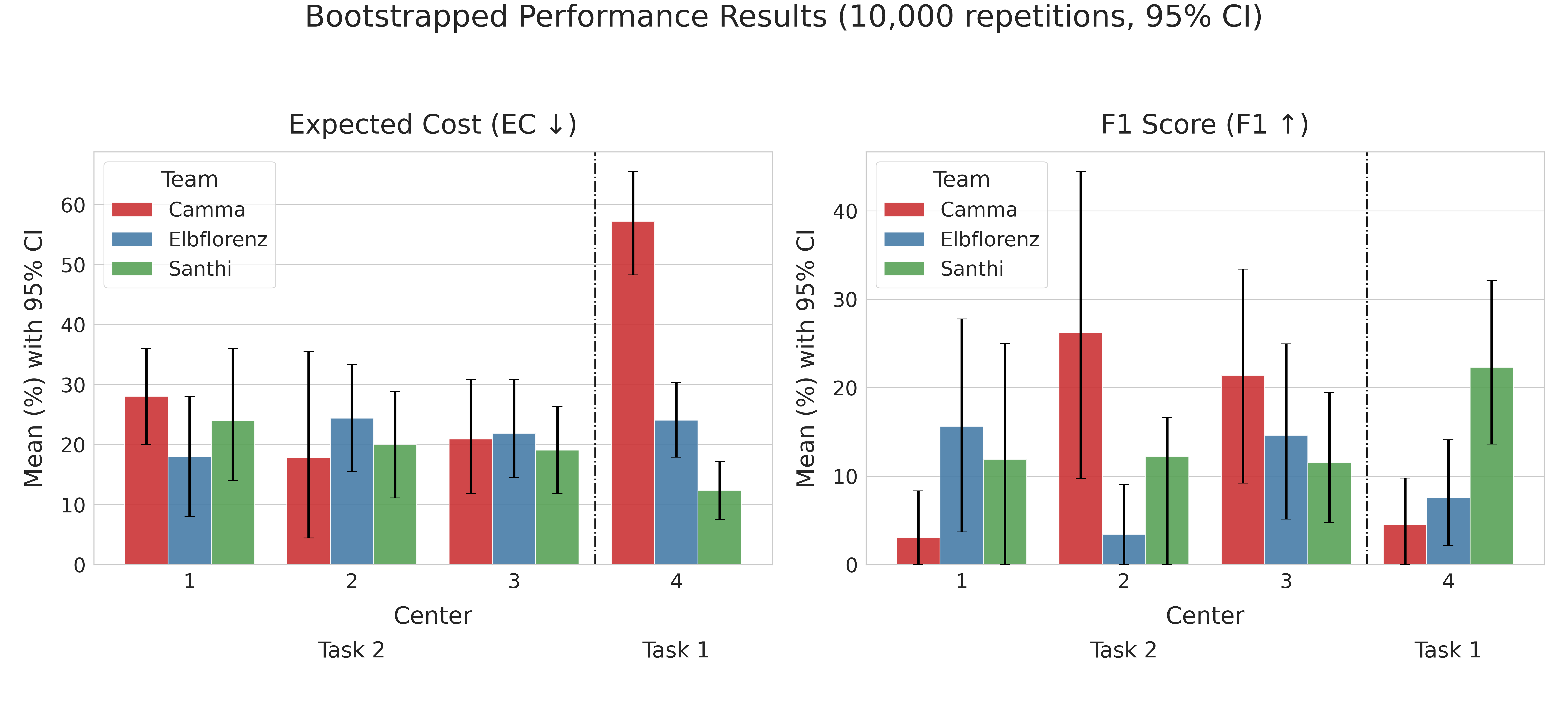}
    \caption{\textbf{Bootstrapped Performance Results.} Expected Cost (EC ↓) and macro F1 (F1 ↑) for the three participating teams, shown as the bootstrap mean with 95\% percentile confidence intervals as error bars, from 10,000 bootstrap repetitions. Centers 1–3 correspond to Task 2 (local adaptation) and Center 4 to Task 1 (generalization). }
    \label{fig:bootstrapping_results}
\end{figure} 

\begin{figure}[htbp]
  \centering
  \begin{subfigure}[b]{0.45\textwidth}
    \includegraphics[width=\linewidth]{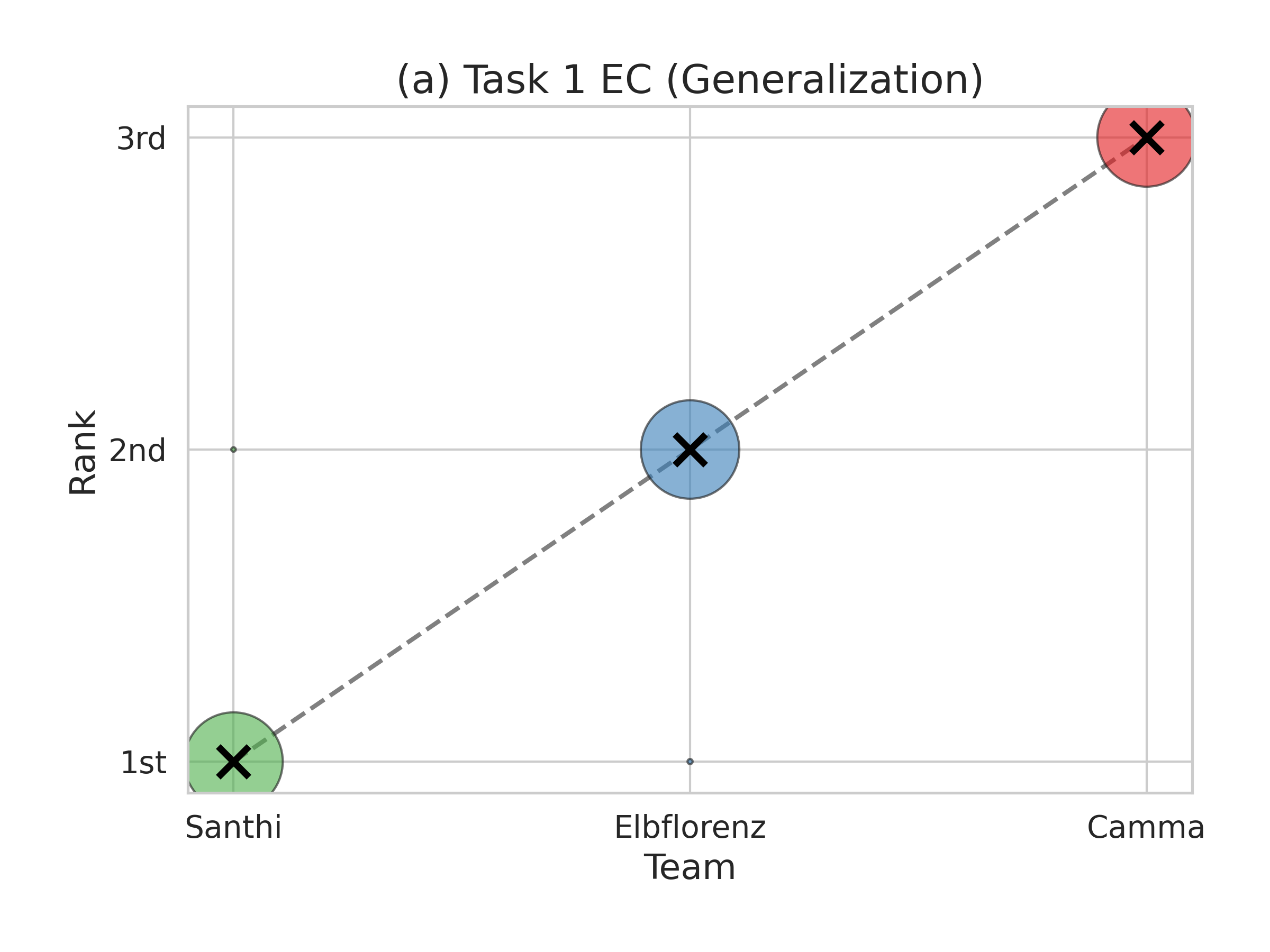}
    \caption{}
    \label{fig:sub1}
  \end{subfigure}
  \hfill
  \begin{subfigure}[b]{0.45\textwidth}
    \includegraphics[width=\linewidth]{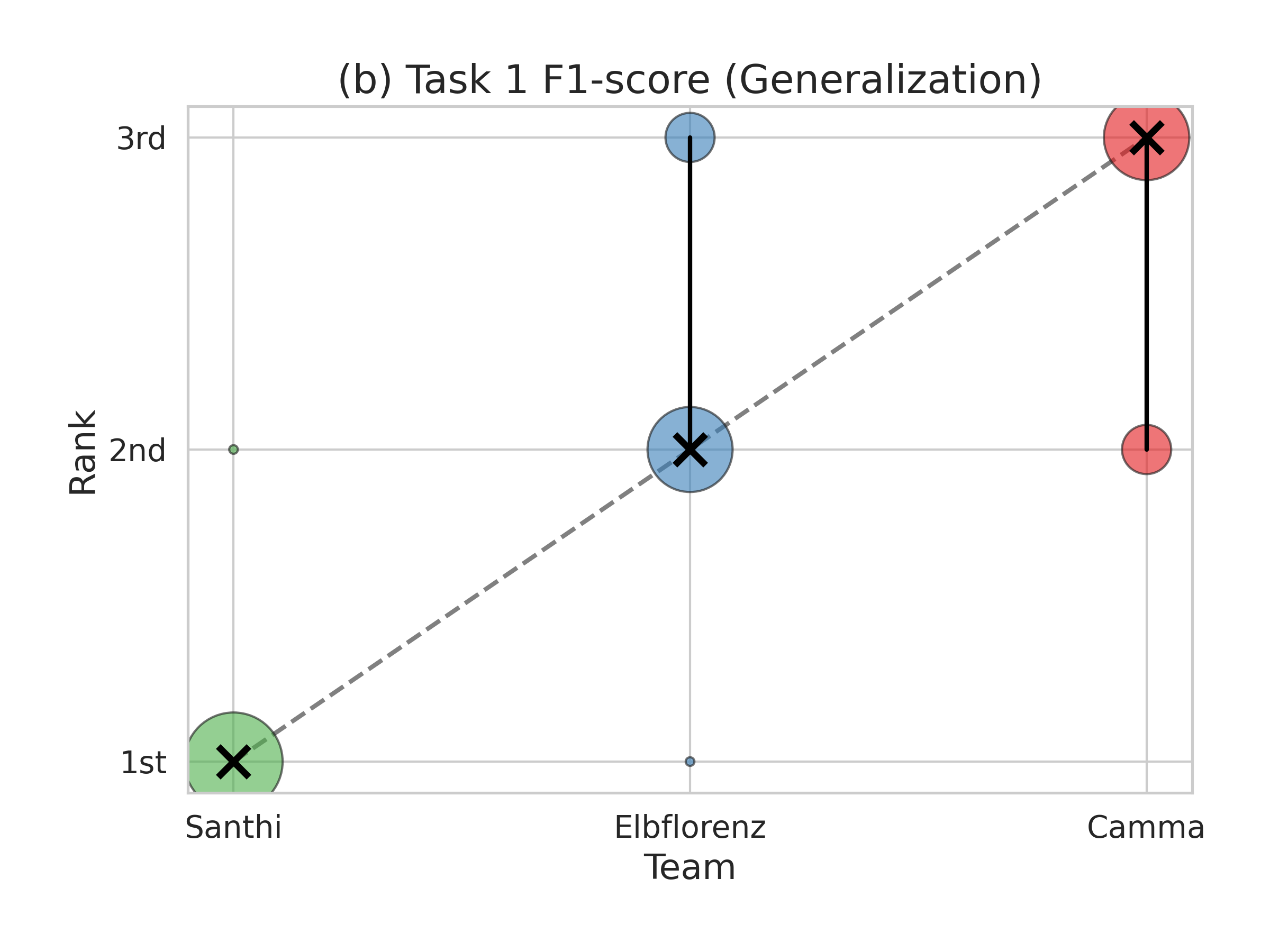}
    \caption{}
    \label{fig:sub2}
  \end{subfigure}
  \vskip\baselineskip
  \begin{subfigure}[b]{0.45\textwidth}
    \includegraphics[width=\linewidth]{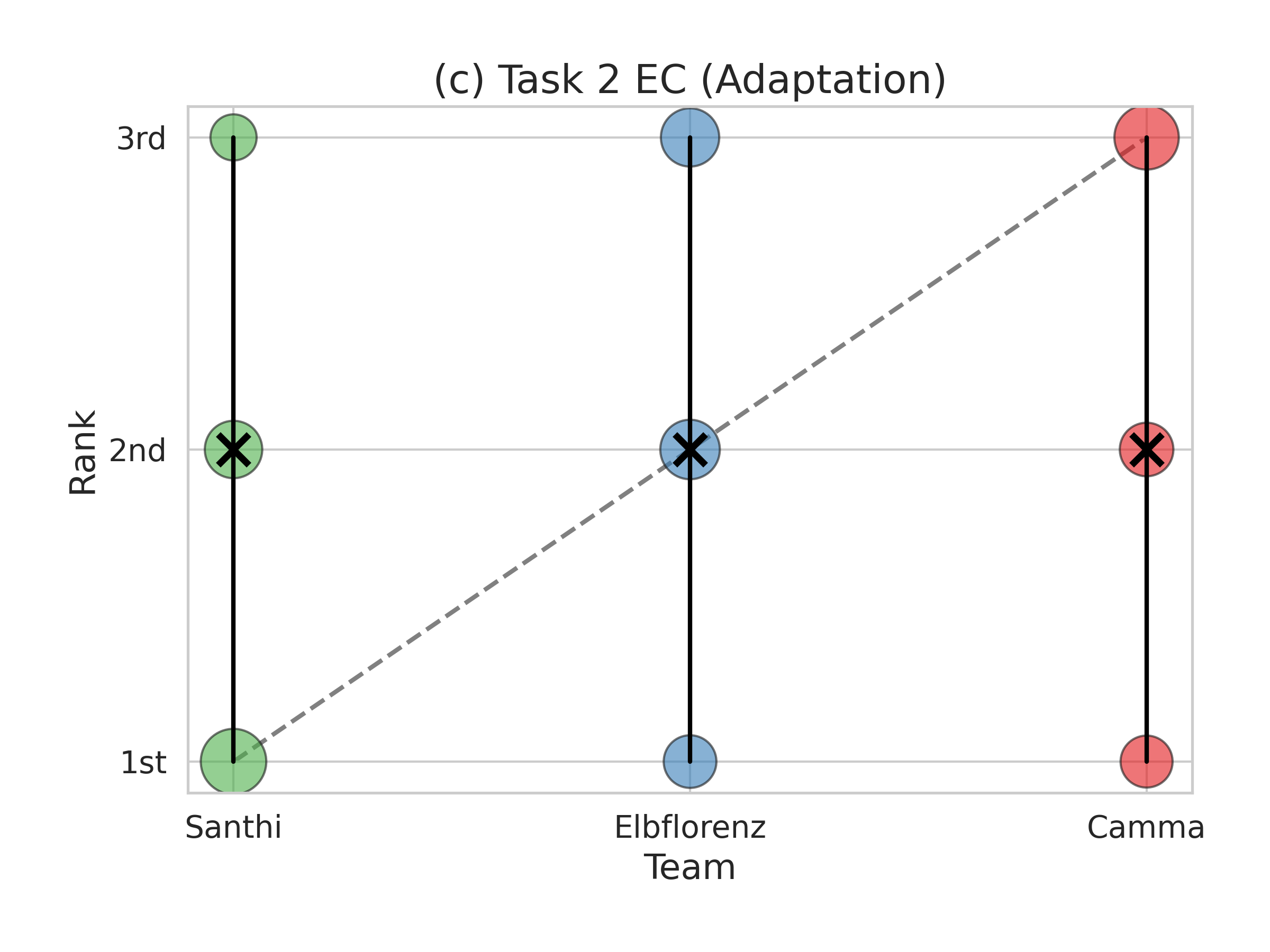}
    \caption{}
    \label{fig:sub3}
  \end{subfigure}
  \hfill
  \begin{subfigure}[b]{0.45\textwidth}
    \includegraphics[width=\linewidth]{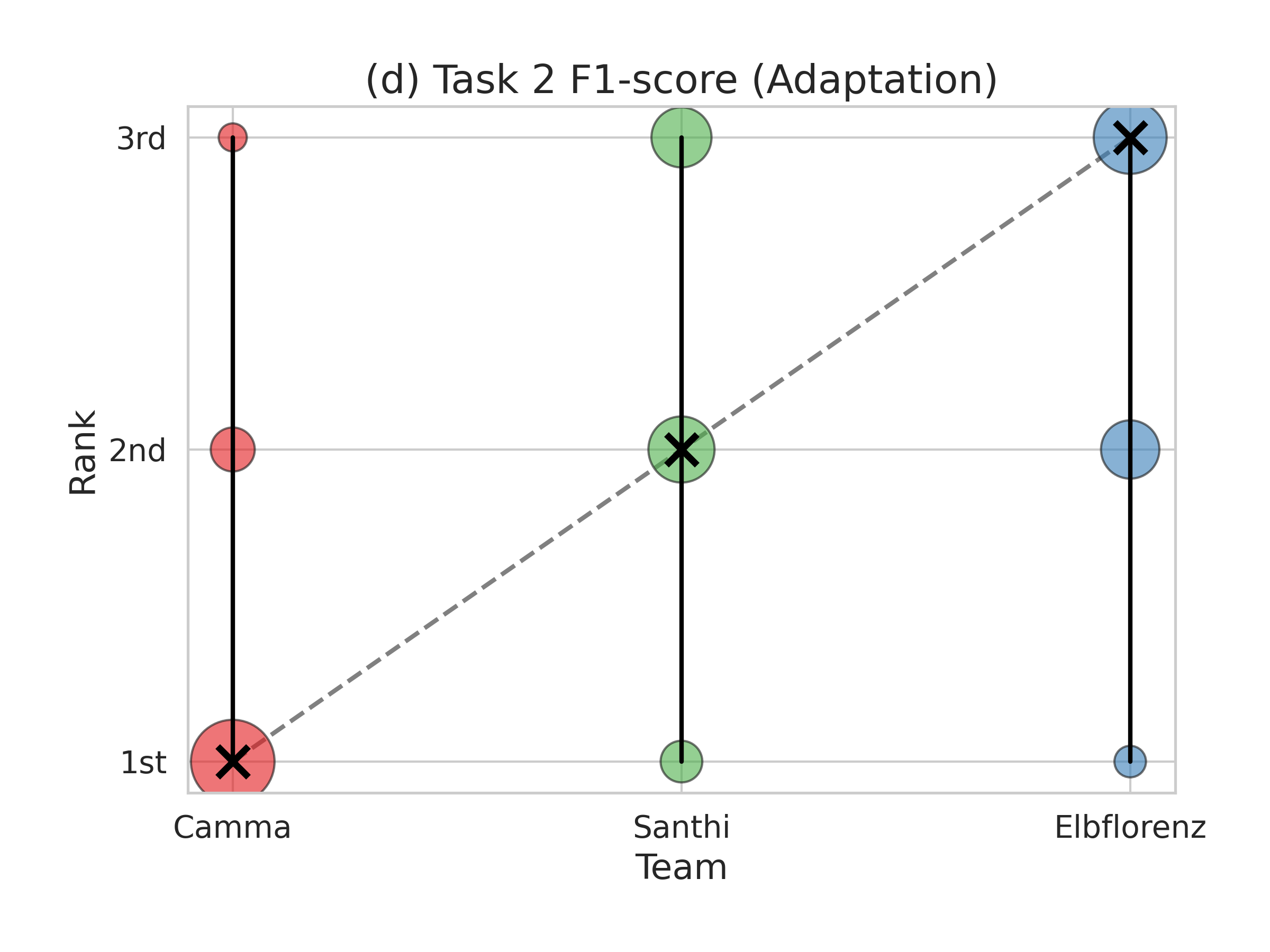}
    \caption{}
    \label{fig:sub4}
  \end{subfigure}
  \caption{\textbf{Ranking Stability.} Bootstrapped ranking distributions for each metric and task, based on 10{,}000 bootstrap iterations. Circle size indicates the percentage of times a team's model achieved a specific rank across samples. Black crosses show median ranks, and black lines denote the 95\% bootstrap confidence intervals. Subfigures (a) and (b) correspond to Task~1 (generalization ability) with metrics EC and F1-score, respectively, while (c) and (d) represent Task~2 (adaptation ability) using the same metrics.}
  \label{fig:rankstability}
\end{figure}

Task 1 assessed the models' ability to generalize to Center 4, which was excluded during training. Overall, performance on this task was modest across all teams (Table \ref{tab:Task1}).

\textbf{Team Santhi} achieved the highest F1-score (23.03\%) and the lowest EC (12.41\%), demonstrating the strongest generalization among the challenge submissions under domain shift. This is evident in the confusion matrix (Figure \ref{fig:confusion_matrix_1}), where Santhi’s predictions tend to align with the diagonal, in contrast to the predictions heavily biased towards a specific class. The model correctly identified class 3 in 8 out of 11 cases but struggled with adjacent stages (e.g., classes 2 and 4) and underrepresented ones (e.g., 0, 1, 5), likely due to class imbalance. Bootstrapped performance metrics (Figure \ref{fig:bootstrapping_results}) indicate that this ordering is consistent under resampling. This is further supported by the ranking stability plots (Figures \ref{fig:sub1} and \ref{fig:sub2}), where Team Santhi ranks first in 99.23\% of bootstrapped samples for F1-score and 99.79\% for EC.

\textbf{Team Camma} obtained the lowest F1-score (4.76\%) and the highest EC (57.24\%). The model predominantly predicted class 0, achieving only three correct predictions across the dataset. This behavior indicates a failure to generalize, likely caused by the model overfitting to the training distribution and being unable to adapt to the significant domain shift in the unseen test set. The poor performance of this collapsed classifier is supported by bootstrapping analysis (Figure \ref{fig:bootstrapping_results}) and by the permutation tests, which separate Camma from every other method on Task~1 EC ($p \le 10^{-4}$, the smallest value attainable with $10{,}000$ permutations). Across 10,000 repetitions, Team Camma consistently ranked as the lowest-performing team in 100\% of cases for the EC and 75.09\% for the F1-score (Figure \ref{fig:rankstability}).

\textbf{Team Elbflorenz} also demonstrated limited performance, predominantly predicting class 2 across inputs (Figure \ref{fig:confusion_matrix_1}). The model achieved an EC of 24.14\% and an F1-score of 7.83\%, ranking slightly above Team Camma. This is likely due to the classifier exhibiting a strong bias towards predicting classes adjacent to the most frequently observed labels in the training data. Confusion matrices and bootstrapped scores confirm weak generalization, placing Elbflorenz in the middle of the three submissions in terms of overall performance (Figure \ref{fig:rankstability}, Figure \ref{fig:bootstrapping_results}).

Paired permutation tests over the 29 Center-4 cases separate the two weakest submissions from the strongest: Camma and Elbflorenz each differ significantly from Santhi on both metrics ($p \le 0.047$), and Camma differs from Elbflorenz on EC ($p = 0.0001$) though not on F1 ($p = 0.592$). The observed differences in generalization under domain shift are therefore resolvable at this sample size for the largest gaps, while the ordering between the two weakest submissions rests on the EC metric alone. The ranking stability plot (Figure \ref{fig:rankstability}) visually reinforces this trend, clearly showing Team Santhi in the lead, followed by Elbflorenz and Camma. The Task 1 component of the leaderboard presented in Table \ref{tab:team-rankings} is therefore well supported.

\subsubsection{Task 2: Center-Specific Adaptation}\label{sec:task2_adaptation}

\begin{table}[htbp]
\centering
\caption{Performance comparison of the three teams on Task 2 across Centers 1, 2, and 3. The table reports Expected Cost (EC, lower is better) and F1-score (higher is better). Best results per metric and center are highlighted in bold.}
\label{tab:Task2}
\begin{tabular}{lccc}
\hline
Team       & Center  & EC \(\downarrow \) & F1 \(\uparrow \)  \\ \hline
Camma      &         & 28.00\% & 3.33\%                                          \\
Elbflorenz & 1       & \textbf{18.00\%}  & \textbf{17.42\%}                               \\
Santhi     &         & 24.00\%          & 14.81\%                             \\ \hline
Camma      &         & \textbf{17.78\%}   & \textbf{30.28\%}                                 \\
Elbflorenz & 2       & 24.44\%            & 3.70\%                                   \\
Santhi     &         & 20.00\%            & 13.33\%                                      \\ \hline
Camma      &         & 20.91\%            & \textbf{22.76\%}                                \\
Elbflorenz & 3       & 21.82\%            & 15.51\%                                  \\
Santhi     &         & \textbf{19.09\%}   & 12.04\%                                      \\ \hline
Camma      &         & 22.23\%            & \textbf{18.79\%}                           \\
Elbflorenz & Average & 21.42\%            & 12.21\%                 \\
Santhi     &         & \textbf{21.03\%}   & 13.40\%                  \\ \hline
\end{tabular}
\end{table}

\begin{figure}[htbp]
    \centering
    \includegraphics[width=\linewidth]{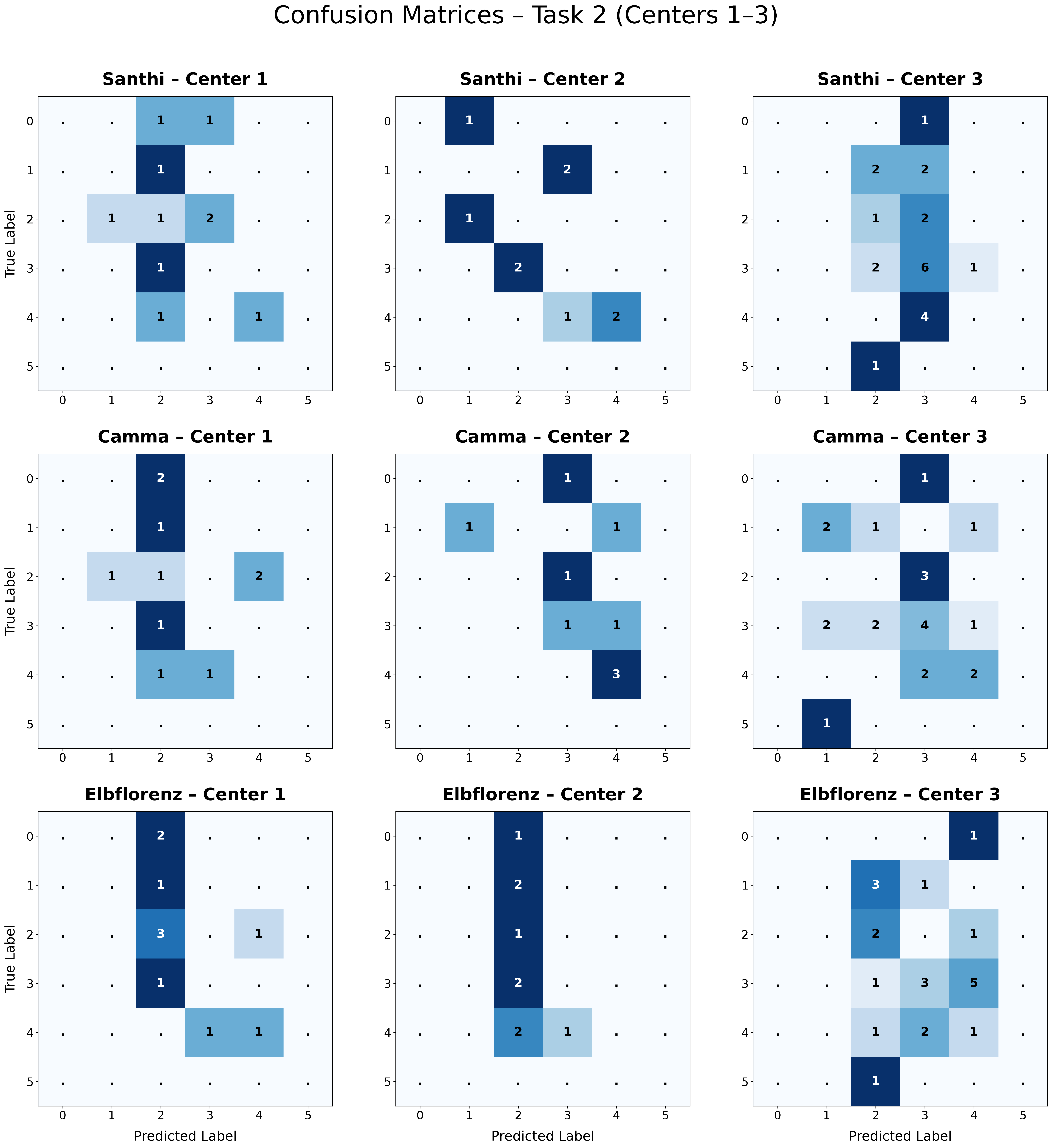}
    \caption{\textbf{Confusion Matrices – Task 2, Centers 1–3.}     
    The values in the confusion matrices are not normalized. The color highlighting is normalized row-wise by true labels.  The diagonal highlights class-wise recall, while off-diagonal values indicate common misclassification patterns.}
    \label{fig:confusion_matrix_2}
\end{figure}

Task 2 evaluated model performance in intra-center settings, where each local model was tested on its dedicated test set from its associated center. Absolute scores are therefore not directly comparable between the two tasks.

\textbf{Team Santhi} returned the narrowest spread of F1-scores across the three centers, at 14.81\% (Center 1), 13.33\% (Center 2) and 12.04\% (Center 3), a range of under three percentage points, and the lowest Task 2 average EC of the three submissions (21.03\%, Table \ref{tab:Task2}). It was not the best submission at any individual center on F1, and it attained the lowest EC at Center 3 only. The confusion matrices in Figure \ref{fig:confusion_matrix_2} show predictions distributed over grades 2 to 4 rather than concentrated on a single grade, with 2 of 10, 2 of 9 and 7 of 22 cases on the diagonal; separation between grades 2, 3 and 4 is limited at all three centers. The bootstrap intervals in Figure \ref{fig:bootstrapping_results} are wide and overlap between centers, and the rank distribution in Figure \ref{fig:rankstability} is correspondingly unstable. For the F1-score, Santhi ranked first in 12.39\% of cases, second in 48.62\%, and third in 38.99\%. Regarding the EC metric, the values were 40.36\%, 36.36\% and 23.28\%.

\textbf{Team Elbflorenz} displayed greater inconsistency across centers, especially in the F1-score (Figure \ref{fig:bootstrapping_results}). The confusion matrices (Figure \ref{fig:confusion_matrix_2}) reveal inconsistent local adaptation: the models for Centers 1 and 2 concentrate their predictions on a single grade, with 7 of 10 and 8 of 9 cases assigned to grade 2, while the model for Center 3 does not separate grades 2, 3 and 4. Notably, the Center 1 model records the highest F1-score of the three submissions at that center (17.42\%) despite this concentration, whereas the same behaviour at Center 2 yields the lowest value of the section (3.70\%). Center 3 is intermediate at 15.51\%. The bootstrap intervals in Figure \ref{fig:bootstrapping_results} indicate that these per-center values are not stable under resampling. For the F1-score, Elbflorenz ranked first in 9.09\% of cases, second in 37.10\%, and third in 53.81\%. Regarding the EC metric, Elbflorenz achieved first place in 36.00\% of cases, second place in 32.30\%, and third place in 31.70\%.

\textbf{Team Camma} shows the widest center-to-center variation of the three submissions (Table \ref{tab:Task2}, Figure \ref{fig:bootstrapping_results}), excelling at Center 2 with an F1-score of 30.28\% and at Center 3 with 22.76\%, but underperforming significantly at Center 1, where the F1-score was 3.33\%. Confusion matrices shown in Figure \ref{fig:confusion_matrix_2} indicate intermediate performance relative to Teams Santhi and Elbflorenz, as the models demonstrate a modest ability to differentiate between classes, evidenced by a visible diagonal tendency. Similar to the other submissions, bootstrapping analyses confirm variability in performance across centers. The ranking for Team Camma also fluctuates across centers, with F1-score rankings of 1st place in 78.52\% of cases, 2nd place in 14.29\%, and 3rd place in 7.19\%. For the EC metric, the rankings were first place in 23.68\%, second place in 31.42\%, and 3rd place in 44.90\%.

The bootstrapped ranking stability analysis (Figures~\ref{fig:sub3} and \ref{fig:sub4}) shows notable fluctuations in team positions within the bootstrapping. While the Task 2 F1-score rankings (Figure~\ref{fig:sub4}) align with the overall standings in Table~\ref{tab:team-rankings}, the EC mean-based rankings (Figure~\ref{fig:sub3}) reveal no clear winner with respect to the median rank. Nevertheless, the frequency of rank positions matches the trend of the general ranking in Table~\ref{tab:team-rankings}. The Task 2 EC ranking is close to uniform, with first-rank probabilities of $0.40$, $0.36$ and $0.24$ for Santhi, Elbflorenz and Camma and mean ranks of $1.83$, $1.96$ and $2.21$. Such variability highlights that Task 2 outcomes are sensitive to small changes in the test set, and claims regarding top-performing submissions on this task are not supported by the evidence available here. Following the criterion of Maier-Hein et al. \cite{maier-hein_why_2018}, no team retained its original EC rank in at least 50\% of bootstrap replicates, and no Task 2 comparison reaches significance under the permutation test. The Task 2 EC ordering therefore carries essentially no information about submission superiority. We report it for completeness as a record of the challenge outcome, not as evidence of relative method quality.

The Wilcoxon signed-rank statistics reported in ~\ref{A:bootstrapping} are computed over the $10{,}000$ bootstrap replicates rather than over the underlying test cases. Their effective sample size is therefore the number of resampling iterations, which is a chosen parameter, and not the $29$ test cases of the held-out center or the $41$ cases underlying the Task 2 average. These values quantify how consistently one method exceeds another under resampling of a fixed test set. They do not constitute inference about a population of centers or patients, and they are reported as stability diagnostics rather than as hypothesis tests, labelled as bootstrap discordance in ~\ref{A:bootstrapping}.

To provide an inferential counterpart, we additionally conducted paired permutation tests over the actual test cases, as described in Section~\ref{sec:ranking}. Five of the six pairwise comparisons on Task 1 reach $p < 0.05$, the exception being Camma versus Elbflorenz on F1 ($p = 0.59$); the significant comparisons involve EC differences of $12$ to $45$ percentage points, or F1 differences of 15 percentage points or more. No Task 2 comparison reaches significance on either metric. The contrast with the bootstrap diagnostics is instructive: comparisons that the resampling diagnostics report as separated at arbitrarily small values are not resolved at the sample size actually available. At $70$ test cases, only very large effects are detectable, and the Task 2 ranking in particular should be interpreted accordingly.

\begin{table}[htbp]
\centering
\caption{Team rankings for the FedSurg Challenge are presented, where lower ranks indicate better performance. The Task 2 EC ranking is not reproducible under resampling and is reported for completeness only; no team retains its position in at least $50\%$ of bootstrap replicates, and no Task 2 comparison reaches significance under the permutation test (Section~\ref{sec:results}). The Final Rank column inherits this instability through the Task 2 average.}
\label{tab:team-rankings}
\renewcommand{\arraystretch}{1.2}
\begin{tabular}{l|cc|c|cc|c|c}
\hline
\multirow{2}{*}{\textbf{Team}} & \multicolumn{3}{c|}{\textbf{Task 1 Ranks}} & \multicolumn{3}{c|}{\textbf{Task 2 Ranks}} & \multirow{2}{*}{\textbf{Final Rank}} \\
                               & EC & F1-score & Avg & EC & F1-score & Avg &                                \\ \hline
Camma                         & 3  & 3        & 3   & 3  & 1        & 2 & 2                            \\
Elbflorenz                    & 2  & 2        & 2   & 2  & 3        & 3  & 2                           \\
Santhi                        & 1  & 1        & 1   & 1  & 2        & 1   & 1                            \\ \hline
\end{tabular}
\end{table}

\subsection{Further Analysis}\label{sec:FurtherAnalysis}
In this section the challenge participant results are compared with the centralized and federated baselines. 
Furthermore, we analyze the communication efficiency of the different approaches. The methods used are described in section~\ref{sec:ranking}.

\subsubsection{Baseline comparison}

\begin{table}[H]
\centering
\small
\caption{Absolute raw test performance comparison including baselines. EC: lower is better; F1: higher is better. Baseline rows in the lower half of the table are contextual references and are not part of the official challenge ranking. Bold numbers mark the best point estimations only.}
\label{tab:baseline_overall}
\begin{tabular}{lcccc}
\hline
	\textbf{Method} & \textbf{Task 1 EC} & \textbf{Task 1 F1} & \textbf{Task 2 Avg EC} & \textbf{Task 2 Avg F1} \\ \hline
Camma      & 57.24\% & 4.76\%  & 22.23\% & 18.79\% \\
Elbflorenz & 24.14\% & 7.83\%  & 21.42\% & 12.21\% \\
Santhi     & 12.41\% & 23.03\% & 21.03\% & 13.40\% \\
 \hline
Cen        & \textbf{10.34\%} & \textbf{26.31\%} & 19.82\% & \textbf{23.51\%} \\
SWARM      & 13.10\% & 19.00\% & 20.41\% & 17.77\% \\
EndoViT+LoRA & 17.24\% & 9.24\%  & 20.13\% & 8.70\% \\
ViViT+LoRA   & 22.76\% & 15.93\% & \textbf{17.60\%} & 21.47\% \\ \hline
\end{tabular}
\end{table}

\begin{table}[H]
\centering
\small
\caption{Center-wise Task~2 baseline results. EC: lower is better; F1: higher is better.}
\label{tab:baseline_centerwise}
\begin{tabular}{lcccccc}
\hline
	\textbf{Baseline} & \multicolumn{2}{c}{\textbf{Center 1}} & \multicolumn{2}{c}{\textbf{Center 2}} & \multicolumn{2}{c}{\textbf{Center 3}} \\ 
                 & EC & F1 & EC & F1 & EC & F1 \\ \hline
Cen   & 24.00\% & 11.43\% & 20.00\% & \textbf{26.43\%} & \textbf{15.45\%} & \textbf{32.67\%} \\
SWARM & 28.00\% & 10.32\% & 17.78\% & 20.95\% & \textbf{15.45\%} & 22.03\% \\ 
EndoViT+LoRA & \textbf{17.78\%} & 10.32\% & 24.44\% & 6.06\%  & 18.18\% & 9.68\% \\
ViViT+LoRA   & 22.22\% & \textbf{12.22\%} & \textbf{13.33\%} & 22.62\% & 17.27\% & 29.57\% \\ \hline
\end{tabular}
\end{table}

To provide contextual reference points, we introduced four baselines. A centralized (Cen) baseline, representing training on pooled data from all centers for Task~1 and on each center's data separately for Task~2, and a SL (SWARM) baseline, applying the same architecture in a decentralized peer-to-peer setup, together provide a matched comparison of pooled and decentralized training under otherwise similar conditions. Two additional PEFT baselines, EndoViT+LoRA and ViViT+LoRA, probe whether parameter-efficient adaptation can serve as a practical alternative to full local fine-tuning. Table~\ref{tab:baseline_overall} summarizes absolute raw test performance across both tasks, while Table~\ref{tab:baseline_centerwise} reports center-wise Task~2 results for all baseline models.

\begin{figure}[htbp]
  \centering
  \begin{subfigure}[b]{0.45\textwidth}
    \includegraphics[width=\linewidth]{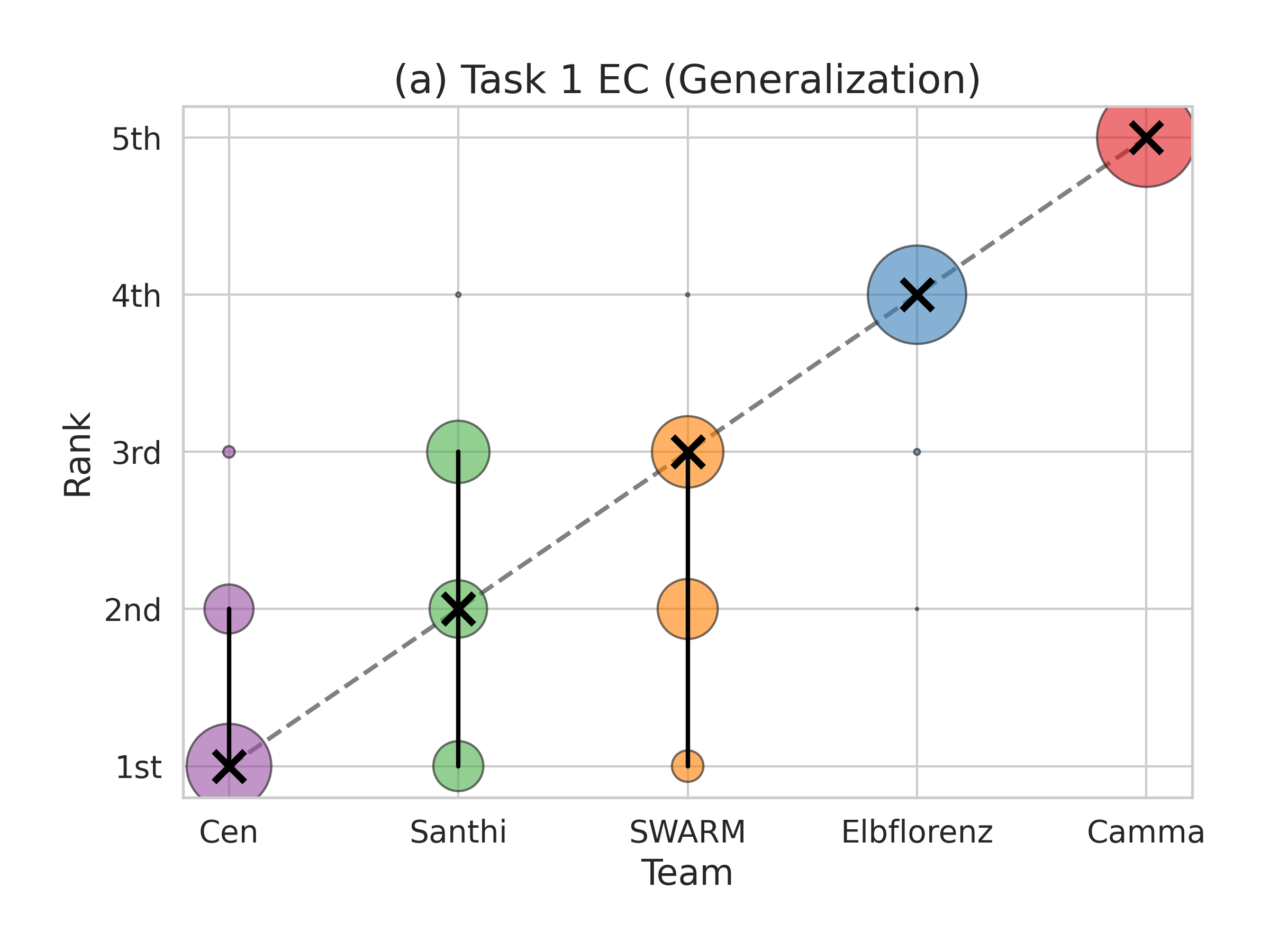}
    \caption{}
    \label{fig:resub1}
  \end{subfigure}
  \hfill
  \begin{subfigure}[b]{0.45\textwidth}
    \includegraphics[width=\linewidth]{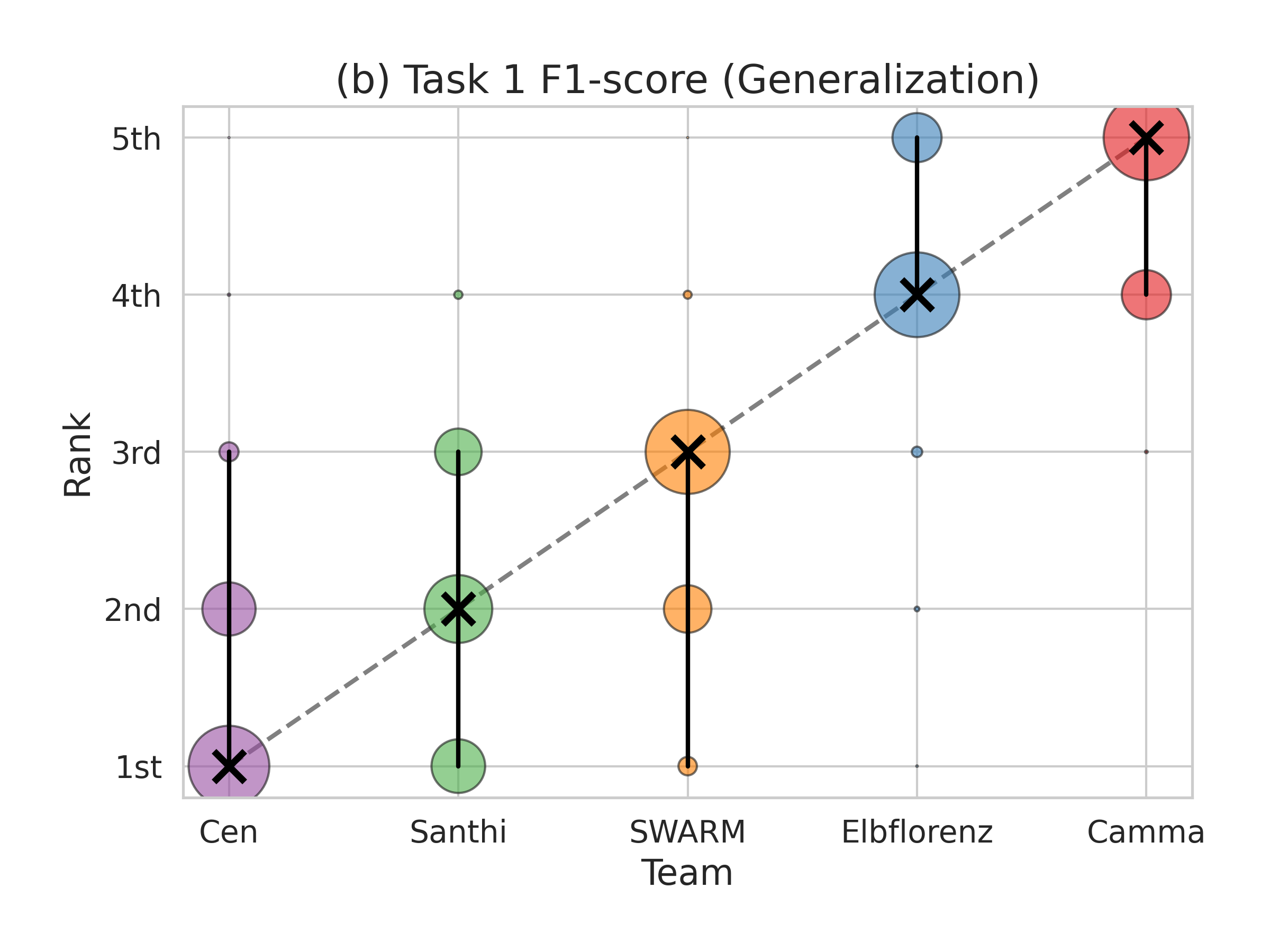}
    \caption{}
    \label{fig:resub2}
  \end{subfigure}
  \vskip\baselineskip
  \begin{subfigure}[b]{0.45\textwidth}
    \includegraphics[width=\linewidth]{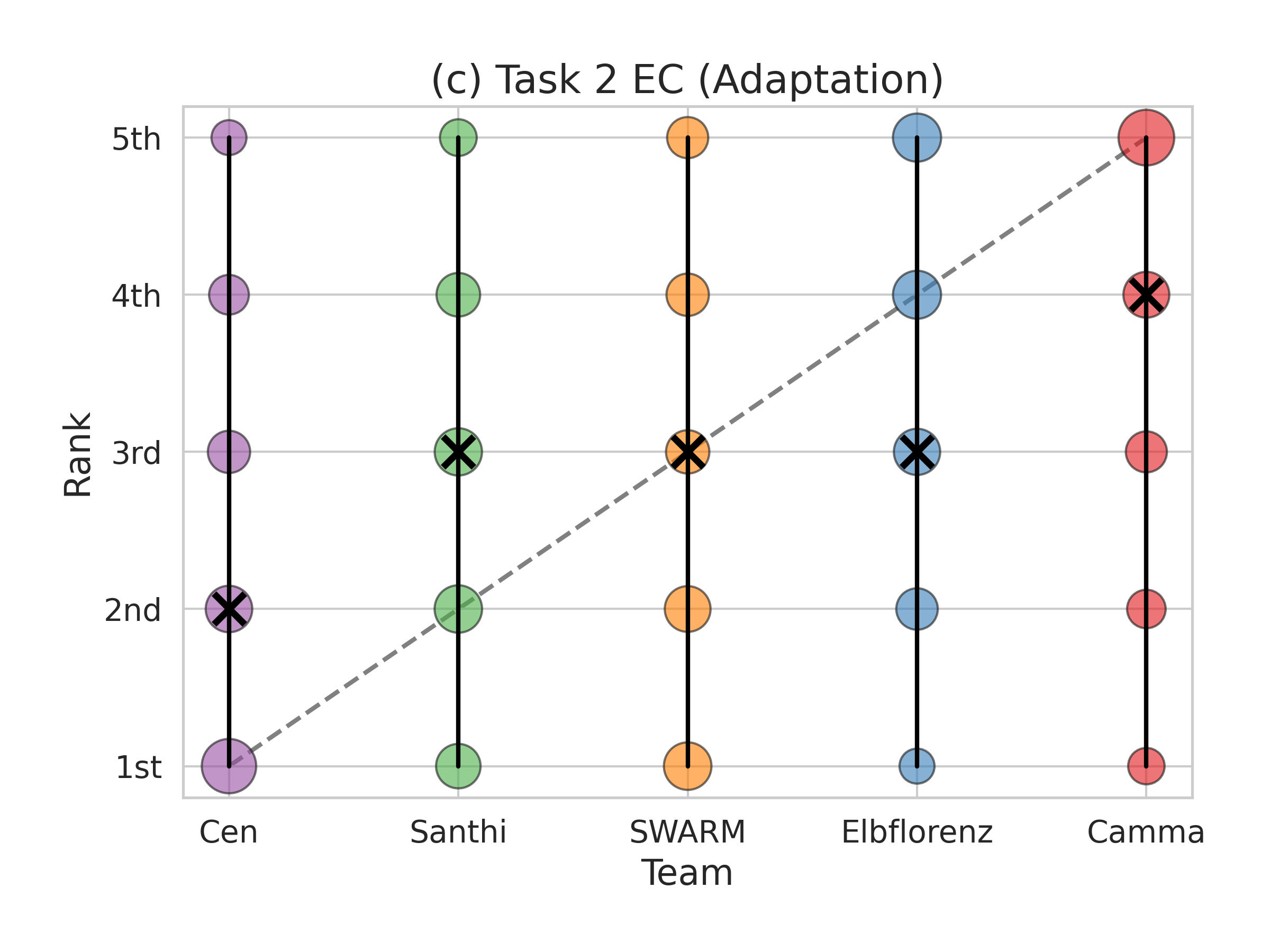}
    \caption{}
    \label{fig:resub3}
  \end{subfigure}
  \hfill
  \begin{subfigure}[b]{0.45\textwidth}
    \includegraphics[width=\linewidth]{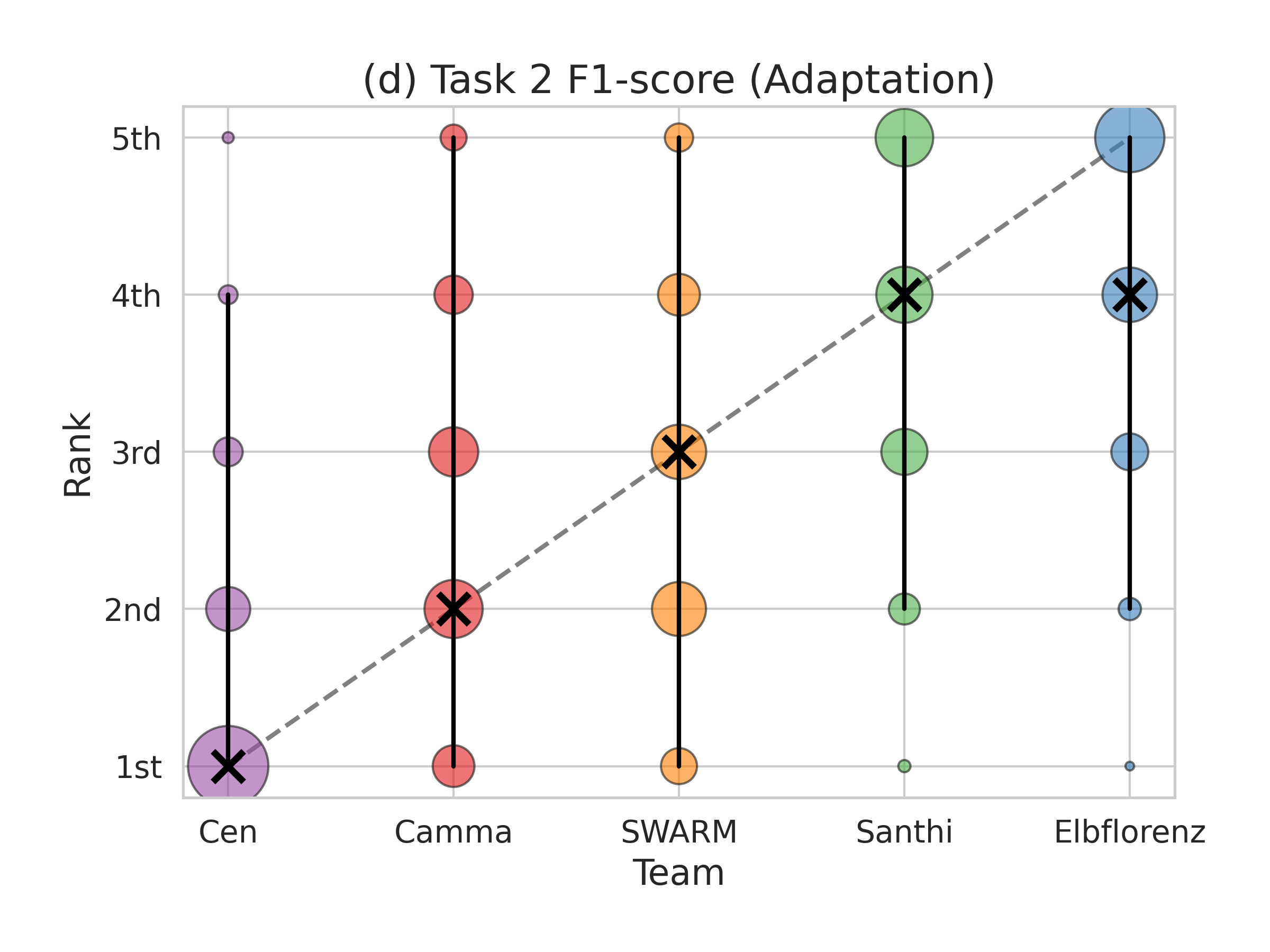}
    \caption{}
    \label{fig:resub4}
  \end{subfigure}
  \caption{\textbf{Ranking Stability.} Bootstrapped ranking distributions for each metric and task, based on 10{,}000 bootstrap iterations. Circle size indicates the percentage of times a team's model achieved a specific rank across samples. Black crosses show median ranks, and black lines denote the 95\% bootstrap confidence intervals. Subfigures (a) and (b) correspond to Task~1 (generalization ability) with metrics EC and F1-score, respectively, while (c) and (d) represent Task~2 (adaptation ability) using the same metrics.}
  \label{fig:re_rankstability}
\end{figure}

\paragraph{Comparative Performance}
The centralized baseline achieved the highest point estimates on both metrics in Task~1 and on F1 in Task~2; on the Task~2 average Expected Cost, ViViT+LoRA (17.60\%) is lower than the centralized baseline (19.82\%). In Task 1 (Generalization), the centralized baseline reached an F1-score of $26.31\%$ and an EC of $10.34\%$, followed by Santhi and the SWARM baseline (compare Figure~\ref{fig:re_rankstability}).

In Task 2 (Adaptation), the performance gap narrowed, reflecting the challenge of center-specific fine-tuning. While the centralized model maintained the lead (Avg. F1: $23.51\%$), the SWARM baseline placed second on the Task 2 average, though its per-center results vary considerably.
Camma and Elbflorenz both improved substantially in Task~2 relative to their Task~1 performance, yet neither reached the performance level of the centralized or SL baselines.

The two PEFT baselines differ substantially in Task~1 F1. EndoViT+LoRA achieved 9.24\% F1 on Task~1 and 8.70\% average F1 on Task~2. Relative to Elbflorenz, the frame-level submission it is derived from, it is the better of the two on three of the four task and metric combinations (Task~1 EC 17.24\% against 24.14\%, Task~1 F1 9.24\% against 7.83\%, Task~2 average EC 20.13\% against 21.42\%) and worse only on the Task~2 average F1 (8.70\% against 12.21\%). Inspection of per-class predictions confirms near-complete collapse to the majority class, and its comparatively low Task~1 Expected Cost should be read in that light: a constant predictor of the pooled training mode attains 13.79\% on the same test set, so a value of 17.24\% reflects concentration on a central grade rather than discrimination. ViViT+LoRA, by contrast, achieved 15.93\% F1 on Task~1 and 21.47\% average F1 on Task~2, ranking second among all methods in Task~2. Notably, at Center~3, ViViT+LoRA reached 29.57\% F1, the highest Task~2 F1 among all non-centralized methods and approaching the centralized baseline (32.67\%). Performance at the smaller centers remained limited (Center~1: 12.22\%, Center~2: 22.62\%). The ordering of the two parameter-efficient baselines is not consistent across metrics: on Task~1 Expected Cost, EndoViT+LoRA (17.24\%) is lower than ViViT+LoRA (22.76\%), reversing the F1 ordering, although both remain above the 13.79\% of a constant predictor. The two backbones differ both in temporal modeling and in pretraining corpus (Section~\ref{sec:baselines}); the attribution of this difference is discussed in Section~\ref{sec:temporal}.

\paragraph{Reference classifiers}
Absolute values on this task are difficult to interpret without a reference point, and macro F1 in particular behaves unintuitively under the observed label distribution. We therefore evaluated four reference classifiers, in two families, on every test set. Two constant classifiers predict a fixed grade: TrivialGlobal predicts the pooled training mode (grade 3), and TrivialLocal predicts each center's own training mode (grades 2, 4 and 3 for Centers 1 to 3 respectively). Two stochastic classifiers were simulated over 10,000 draws: RandomUniform samples all six grades with equal probability, and RandomPrior samples according to training set prevalence. All references are derived from training labels only and are excluded from every rank column. Table ~\ref{tab:reference_classifiers} reports their values.

Four observations follow.

First, the attainable macro F1 is bounded below 100\% wherever a grade is absent from the test set, since absent classes contribute zero against a fixed denominator of six. The ceiling is 83.33\% at Centers 1, 2 and 4, and 88.89\% for the Task 2 average. Reported F1 values should be read against this bound rather than against unity. At the other end of the scale, the most generous chance model is a stochastic classifier matching the \emph{test} prevalence exactly, for which each class contributes $\text{F1}_c = p_c$ in expectation and the macro average is therefore $1/C$, that is $16.67\%$ for six grades. That model presupposes knowledge of the test distribution and is not attainable in practice; the two stochastic references reported here use training prevalence or a uniform draw and consequently sit lower, at $14.30\%$ and $11.91\%$ on the held-out center.

Second, on Task 1 the constant classifier reaches 13.79\% EC, which two of the three submissions fail to improve upon (Elbflorenz 24.14\%, Camma 57.24\%); both parameter-efficient baselines also fail to improve on it (EndoViT+LoRA 17.24\%, ViViT+LoRA 22.76\%). Camma's cost exceeds even uniform random guessing (33.55\%), reflecting a predicted distribution systematically displaced from the ordinal position of the true labels.

Third, at Center 1 the constant classifier predicting that center's training mode reaches 20.00\% EC, which five of the seven evaluated methods fail to improve upon, including the centralized baseline (24.00\%) and the Swarm baseline (28.00\%). Only Elbflorenz (18.00\%) and EndoViT+LoRA (17.78\%) are more accurate under the ordinal cost, and both by at most 2.2 percentage points. On macro F1 the same center places both baselines below the prior matched chance level. Center 1 is therefore a setting in which neither centralized nor decentralized training establishes a meaningful advantage over trivial prediction.

Fourth, at Center 3 no submission improves on the constant classifier on EC (Santhi 19.09\%, Camma 20.91\%, Elbflorenz 21.82\%, against 18.18\%), while the centralized, Swarm and ViViT+LoRA baselines do (15.45\%, 15.45\% and 17.27\%); EndoViT+LoRA matches the reference exactly at 18.18\%. Center 3 contributed 80 of 153 training videos, so the submissions fail to exceed a constant predictor at precisely the node where local data is most abundant.

Across all test sets and both metrics, eleven method, test set and metric combinations fall below at least one chance classifier, giving twenty individual comparisons, since a result may fall below both stochastic references. On the Task 2 average F1, Elbflorenz (12.21\%) and Santhi (13.40\%) both sit below the prior matched chance level (14.65\%). These comparisons are reported in full in the table~\ref{tab:reference_comparisons}.

The constant references require one caveat. TrivialLocal at Center 2 predicts grade 4, which lies near the edge of the ordinal scale, so the linear distance penalty inflates its cost to 31.11\% EC against 24.44\% for the mid scale TrivialGlobal on identical data. Comparisons at Center 2 are therefore made against TrivialGlobal, and the fact that all methods improve on TrivialLocal there should not be read as evidence of adaptation quality.

\paragraph{Rank Stability and Statistical Significance}
To address the stability of these findings beyond point estimates, we conducted a stability analysis using $10,000$-iteration bootstrapping. This process quantifies the probability of each team achieving a specific rank under different data distributions, as visualized in Figure~\ref{fig:re_rankstability}.
As shown in Figure~\ref{fig:re_rankstability}(a) and (b), the rankings for Task 1 were comparatively stable. For the F1-score, the centralized baseline retained the 1st rank in $65.7\%$ of iterations, while SWARM maintained the 3rd rank in $69.8\%$ of cases. In contrast, Task 2 exhibited higher volatility (Figure~\ref{fig:re_rankstability}(c) and (d)), particularly in the EC metric. No method retained its nominal Task 2 EC rank in more than $37.9\%$ of replicates, including the centralized baseline, and Camma attained the first rank in $8.6\%$ of replicates despite ranking last on point estimates. Restricted to the three submissions alone, the Task 2 EC ranking is close to uniform, with 1st-rank probabilities of $0.40$, $0.36$ and $0.24$ for Santhi, Elbflorenz and Camma and mean ranks of $1.83$, $1.96$ and $2.21$. Following the criterion of Maier-Hein et al.~\cite{maier-hein_why_2018}, the Task 2 EC ordering carries essentially no information about submission superiority.

As set out in Section~\ref{sec:task2_adaptation}, the Wilcoxon signed-rank statistics in \ref{A:bootstrapping} are computed over bootstrap replicates rather than over test cases and are reported as stability diagnostics rather than as hypothesis tests. They apply here in the same way, with the baselines included.

To provide an inferential counterpart, we additionally conducted paired permutation tests over the actual test cases (Section~\ref{sec:ranking}). Twelve of forty pairwise comparisons reach $p < 0.05$, all of them on Task 1 and all involving Camma or Elbflorenz, whose Task 1 EC differences from the remaining methods range from $11$ to $47$ percentage points. No Task 2 comparison reaches significance on either metric (minimum $p = 0.507$ for EC and $0.073$ for F1). In particular, the Santhi submission and the SWARM baseline achieve Task 2 average Expected Cost values of $21.03\%$ and $20.41\%$, a difference of $0.62$ percentage points, with heavily overlapping bootstrap intervals ($16.00$--$26.22\%$ and $14.46$--$26.77\%$), a bootstrap win probability of $0.407$, and a permutation $p$-value of $0.844$; their adaptation performance under the ordinal cost is statistically indistinguishable at this sample size, notwithstanding their divergence in F1-score. We note that this conclusion rests on the width of the intervals relative to the effect rather than on the point estimates being close. At $70$ test cases, only very large effects are resolvable, and the ranking should be interpreted accordingly.

\begin{figure}[htbp]
    \centering
    \includegraphics[width=\linewidth]{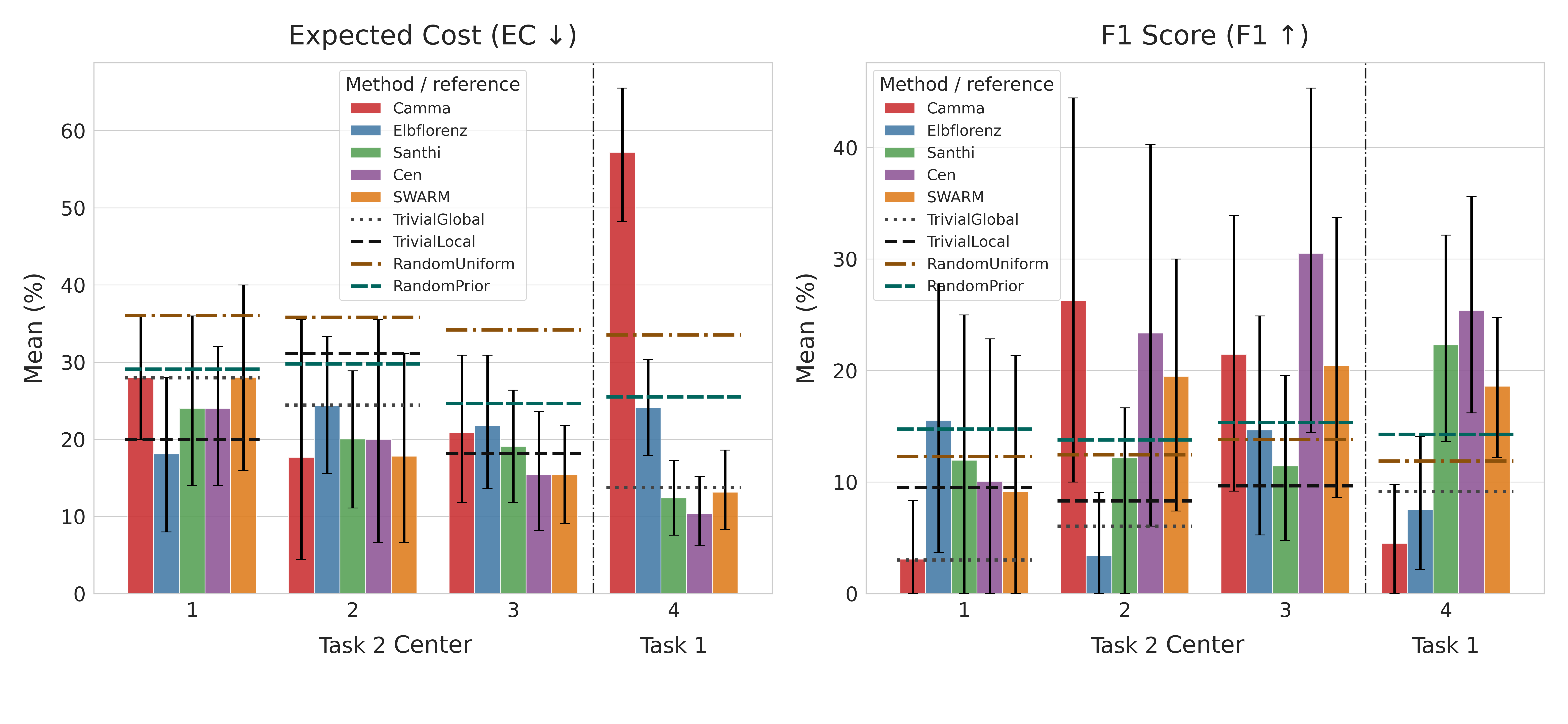}
    \caption{\textbf{Bootstrapped Performance Results with Baselines.} Expected Cost (EC $\downarrow$) and macro F1 (F1 $\uparrow$) for the three participating teams and baselines, shown as the bootstrap mean with 95\% percentile confidence intervals as error bars, from 10,000 bootstrap repetitions. Centers 1--3 correspond to Task 2 (local adaptation) and Center 4 to Task 1 (generalization). Horizontal lines mark four non-ranked reference classifiers derived from training labels only: TrivialGlobal, a constant predictor of the pooled training mode (grade 3); TrivialLocal, a constant predictor of each center's own training mode (grades 2, 4 and 3 for Centers 1 to 3); RandomUniform, sampling all six grades with equal probability; and RandomPrior, sampling according to training set prevalence. The two stochastic references are simulated over 10,000 draws and shown as their mean. The two constant references coincide at Center 3, where the local and pooled training modes are identical, and TrivialLocal is undefined for Task 1, whose held-out center contributes no training labels. Reference values are contextual only and are excluded from all rank computations.}
    \label{fig:re_bootstrapping_results_w_baseline}
\end{figure}

The variability of these outcomes is further illustrated in Figure~\ref{fig:re_bootstrapping_results_w_baseline}, where error bars represent the $95\%$ percentile confidence interval across bootstrap replicates. The narrower distributions for Task 1 compared to the wider variance in Task 2 underscore the inherent difficulty of the adaptation task, where local center characteristics significantly influence model stability.

\subsubsection{Communication efficiency analysis}

Communication cost is a critical metric in decentralized machine learning because the frequent transmission of high-dimensional model parameters, weights, or gradients over limited or unstable network bandwidth can become a significant bottleneck~\cite{mcmahan_communication-efficient_2017}. In many medical AI applications, the time required for network synchronization often dwarfs the time spent on local GPU computation. Consequently, optimizing these costs is essential for making collaborative learning scalable and practical for real-world clinical scenarios.

In a traditional FL architecture, the total communication costs can be calculated using the following formula:

\begin{equation}
C_{\text{total}} = 2 \times K \times T \times |W|
\end{equation}

Where:
\begin{itemize}
    \item $C_{\text{total}}$ is the total communication cost (data transferred).
    \item The factor of $2$ accounts for the bi-directional transfer (upload and download) of model parameters between the client and the central server in each round.
    \item $K$ is the number of participating clients (or nodes) per round.
    \item $T$ is the total number of communication rounds.
    \item $|W|$ represents the size of the model parameters or weights.
\end{itemize}

However, for SL, the communication cost is calculated differently due to its peer-to-peer (P2P) nature. Since a central server does not exist, nodes synchronize weights directly through a decentralized mesh. In a leader-based synchronization (as utilized in the SurgTempoNet framework), one node is elected to aggregate weights from its peers. Consequently, the communication cost formula is adjusted to reflect the $K-1$ active connections:

\begin{equation}
C_{\text{total}} = 2 \times (K - 1) \times S \times |W| 
\end{equation}

Where $S$ represents the total number of synchronization events across the entire training duration. In the SurgTempoNet implementation, synchronization is governed by a frequency parameter rather than fixed rounds. The total number of synchronization events $S$ for a node is defined as:

\begin{equation}
    S = \sum_{n=i}^{K} \left\lfloor \frac{E \times B_i}{f_{\text{sync}}} \right\rfloor + 1
\end{equation}

Where:
\begin{itemize}
    \item $E$ is the number of local epochs performed.
    \item $B_i$ is the number of batches contained in each epoch for center $i$.
    \item $f_{\text{sync}}$ is the synchronization frequency (e.g., every 100 batches).
    \item The $+1$ accounts for the initial synchronization at the start of training.
\end{itemize}

By utilizing $2(K-1)$, the model accounts for the fact that the aggregating node does not incur external network traffic for its own local weights. This provides a more rigorous representation of the physical network traffic generated during the decentralized training of surgical video classifiers across heterogeneous hospital sites.

Table \ref{tab:communication} reveals again significant variance in both architectural complexity and the resulting communication overhead during the FL process.

\paragraph{Model Architecture and Paramerization}
The models utilized by Santhi, Elbflorenz, and our SL Baseline share a similar scale, ranging around 88M parameters.
In contrast, team Camma implemented a significantly more compact architecture. With only 24M parameters, Camma's model is approximately 73 \% smaller than the other methods.
All models utilize 32-bit floating-point precision, resulting in a model size of approximately 340 MB for the larger architectures and around 92 MB for Camma's compact model.

\paragraph{Synchronization Events and Total Communication Costs}
As discussed, the total communication cost is a function of both the model size and the frequency of synchronisation events.
The SL Baseline, with its high synchronization frequency (100 events), incurs the highest communication cost, totaling approximately 133.14 GB across all clients and rounds, despite saving communication compared to the other methods by using a P2P approach.
Furthermore, the impact of synchronization frequency is most evident when comparing Santhi and Elbflorenz.

Team Elbflorenz, with 50 synchronization events, results in a total cost of around 97.47 GB.
Team Santhi, with only 20 synchronization events, significantly reduces the communication overhead to approximately 39.63 GB. 

Finally, team Camma, with the smallest model and lowest synchronization frequency (10 events), achieves the lowest communication cost of about 5.38 GB.

In summary, team Camma emerged as the most communication-efficient team by a wide margin. By combining a reduced parameter count with a highly conservative synchronization strategy (10 sync events), they achieved a total communication cost of only 5.38 GB. Compared to the SL Baseline, this represents a $95.9\%$ reduction in network traffic, highlighting the efficacy of aggressive model compression and sparse synchronization in federated environments.

In the current implementation, both PEFT baselines transmitted full model snapshots during aggregation, incurring communication costs comparable to full-model FL methods. Since only the LoRA adapter weights differ between federation rounds ($\sim$940--969K parameters, $\sim$3.8~MB per model), a dedicated LoRA-only synchronization strategy would reduce per-round transmission to less than 2\% of the costs reported in Table~\ref{tab:communication}. This efficiency advantage was not exploited here, but represents a compelling practical benefit of PEFT-based FL for future implementations.

\begin{table}[]
\centering
\caption{Summary of architectural parameters and cumulative synchronization data costs per team.}
\label{tab:communication}
\setlength{\tabcolsep}{2pt}
\renewcommand{\arraystretch}{1.1}
\begin{tabularx}{\linewidth}{@{}>{\raggedright\arraybackslash}p{2.35cm}|>{\raggedright\arraybackslash}X>{\raggedright\arraybackslash}X>{\raggedright\arraybackslash}X>{\raggedright\arraybackslash}X@{}}
\hline
\textbf{Team}               & \textbf{SL Baseline}          & \textbf{Santhi}             & \textbf{Elbflorenz}          & \textbf{Camma}            \\ \hline
\textbf{Parameters}         & 89,351,470                    & 88,651,014                  & 87,213,038                   & 24,086,014                \\
\textbf{Data Type}          & float32 (4 bytes/param)       & float32 (4 bytes/param)     & float32 (4 bytes/param)      & float32 (4 bytes/param)   \\
\textbf{Model size}         & 357,405,880 B (340.85 MB)     & 354,604,056 B (338.18 MB)   & 348,852,152 B (332.69 MB)    & 96,344,056 B (91.88 MB)   \\
\textbf{Clients}            & 3                             & 3                           & 3                            & 3                         \\
\textbf{Sync Events}        & 100                           & 20                          & 50                           & 10                        \\
\textbf{Total Costs}        & 142,962,352,000 B (133.14 GB) & 42,552,486,720 B (39.63 GB) & 104,655,645,600 B (97.47 GB) & 5,780,643,360 B (5.38 GB) \\ \hline
\end{tabularx}
\end{table}

\section{Discussion}

The FedSurg challenge should be understood as a proof-of-concept benchmark: the appendectomy task was selected for the combination of a standardised workflow, an established grading system, and genuine multi-center availability, in a procedure not previously represented in surgical vision datasets~\cite{kolbinger_appendix300_2025}. The following discussion focuses on methodological insights transferable to future FL applications on surgical video, structured around two central questions: first, what architectural and algorithmic choices best support global generalization to unseen centers; and second, whether and how local adaptation can be reliably achieved without sacrificing the globally learned representation.

\subsection{The Role of Temporal Modeling}\label{sec:temporal}
A key finding of this challenge is that temporal modeling was the design choice most clearly associated with performance in this heterogeneous setting. Frame-level approaches, whether operating through majority voting (Team Elbflorenz) or embedding-based prototype comparison (Team Camma), are inherently limited by the high intra-video visual variance present in laparoscopic recordings. Differences in laparoscopic hardware, lighting conditions, and patient anatomy render individual frames unreliable predictors of patient-level inflammation grade. The metric learning approach of Team Camma illustrates this most clearly: while it achieved competitive performance at centers whose visual characteristics were represented in training, it failed to map embeddings of the unseen center to established training prototypes, revealing the brittleness of a feature space built on individual frames.

Video-level temporal models, by contrast, capture the progression of visual features across the surgical sequence, providing contextual information no single frame can supply. The ViViT-based submission of Team Santhi, processing 32 frames jointly, consistently achieved the strongest generalization to the unseen center in Task 1, outperforming both frame-level approaches despite using fewer communication resources. This finding is consistent with prior work on decentralized surgical video analysis using the full Appendix300 cohort, of which the challenge data is a subset \cite{benchmarking_paper}, in which temporal architectures similarly outperformed frame-level methods across patient-level prediction tasks. The two parameter-efficient baselines come closest to a controlled comparison in this work. EndoViT+LoRA and ViViT+LoRA hold the aggregation strategy, optimizer, learning-rate schedule, round count and classification head fixed, and differ in the frozen backbone. That backbone differs in two respects, frame-level versus temporal processing and endoscopic versus general-domain pretraining, so the resulting gap on Task~1 (F1 9.24\% versus 15.93\%) does not isolate temporal modeling as the sole cause. It is consistent with the ordering observed among the submissions, and the fact that the domain-specific backbone is the weaker of the two argues against pretraining domain as the explanation. The gap is also metric-dependent. On Task~1 Expected Cost the ordering reverses, with the frame-level EndoViT+LoRA at 17.24\% against 22.76\% for ViViT+LoRA, because collapse onto a central grade is inexpensive under a linear ordinal penalty even when it carries no discriminative information; neither improves on the 13.79\% of a constant predictor there. In the controlled comparison the temporal advantage therefore holds on three of the four task and metric combinations rather than on all four, and the claim that temporal modeling is the dominant architectural factor rests primarily on the submissions, where the ordering is consistent across both metrics on Task~1.

With respect to the generalization vs. adaptation trade-off, temporal modeling also offered a partial advantage in Task 2: Team Santhi's frozen ViViT backbone preserved global spatiotemporal representations during local fine-tuning, limiting the degree to which center-specific updates could corrupt the globally learned feature space. This suggests that architecture choice is not only a generalization strategy but also a safeguard against the instability of local adaptation. This reading requires one qualification: Santhi's Task~2 average F1 of 13.40\% falls below the prevalence-matched chance level of 14.65\% (Section~\ref{sec:FurtherAnalysis}), so the advantage is visible in the consistency of the Expected Cost across centers rather than in absolute discriminative performance, and no Task~2 comparison is statistically significant in any case. This point is revisited in Section~\ref{sec:personalized_fl}.

\subsection{Dataset Scale as the Binding Constraint}\label{sec:datascale_discussion}
The generally low absolute scores across all submissions raise an important question: how much of the low performance is attributable to the task and data regime rather than to federation? The centralized baseline provides a partial answer. Even when pooling all multi-institutional data onto a single server, thereby removing all federated constraints, the centralized model achieved only 26.31\% F1-score on the unseen center. The underlying task of patient-level appendicitis grading from surgical video is therefore intrinsically difficult, regardless of the learning paradigm employed, due to the high visual variance across centers and the inherent ambiguity of fine-grained clinical grading.

The reference classifiers reported in Section~\ref{sec:FurtherAnalysis} sharpen this picture. On the unseen center both baselines exceed all reference classifiers on point estimates, but the margins differ in kind. Under macro F1 the centralized model sits 2.14 standard deviations above the prevalence-matched chance level, a value reached by only 3.19\% of chance draws, whereas the decentralized baseline sits 0.80 standard deviations above it and is matched or exceeded by 18.25\% of draws; on this metric it is therefore above chance descriptively but not distinguishable from it. Under the EC both baselines fall far below the chance references, and the binding comparison is instead the constant classifier, which they improve on by 3.45 and 0.69 percentage points respectively. The two metrics behave differently because macro F1 is a weak instrument here: absent classes score zero against a fixed denominator of six, and the metric ignores how far a prediction lies from the true grade. The EC, which penalizes ordinal distance, retains signal that macro F1 discards, and should be treated as the primary indicator of model quality throughout.

This does not establish that decentralization causes the gap. The two baselines are not statistically separable from one another at this sample size, and we therefore make no claim of a quantified decentralization penalty. What the comparison does show is that the two paradigms sit on opposite sides of the chance threshold on one metric, and that the margin over trivial prediction is narrow for both. This distinction should be drawn carefully: that one baseline separates from chance while the other does not is a statement about two separate contrasts, not about the contrast between them. The direct comparison of the two is not significant on either metric ($p = 0.243$ for F1 and $p = 0.357$ for EC, Table~\ref{tab:permutation}), and a difference in separability from a common reference is not evidence of a difference between the methods themselves.

The adaptation setting is less favorable still. At Center~1 only two of the seven evaluated methods improve on a constant classifier, and the centralized baseline is not among them; at Center~3 no participant submission improves on it, while three of the four organizer baselines do. Aggregate Task~2 scores therefore conceal substantial variation between centers, and center-level results should be read alongside the averages rather than replaced by them.

The most plausible explanation is data scale. With 153 training videos across three training centers, federated training fragments an already small pool, leaving individual nodes with too few samples to learn robust local representations. A subsequent study applying a related decentralized pipeline to a larger multinational cohort reports performance comparable to centralized training \cite{benchmarking_paper}; that work differs in cohort size, center count and task definition, so it is not a controlled comparison, but the contrast points to the data regime rather than the decentralization mechanism as the binding constraint. Scaling the dataset, in both total videos and contributing centers, is a necessary condition for a more definitive evaluation of FL on surgical video.

\subsection{Data Heterogeneity and the Limits of Local Adaptation}~\label{sec:heterogeneity}
The challenge results highlight that non-IID data remains the central unsolved problem in federated surgical video analysis. Distribution shift across centers originates from two compounding sources: hardware-induced feature skew, including differences in laparoscopic systems, resolution, lighting, and cropping, and patient population heterogeneity, including differences in age distribution, disease severity prevalence, and surgical practice between academic and community centers. Current aggregation strategies, including FedAvg and FedSAM, partially mitigate but do not resolve these shifts.

These effects are further amplified by inter-center data quantity skew. Center 3 contributed 80 of the 153 training videos, causing global models to disproportionately reflect Center 3's data distribution. The bootstrapped rank stability analysis illustrates the consequence: Task 1 rankings were comparatively stable, and the permutation tests separate the frame-level submissions from the temporally-informed ones, though not the leading methods from one another; Task 2 rankings showed pronounced instability, with no submission retaining its rank on the EC metric in more than 50\% of bootstrap replicates.

Degenerate prediction behavior is present across submissions, but its distribution differs from what a purely fine tuning based account would predict. The most pronounced collapse occurs on Task 1, where the global model is evaluated on an unseen center without any local adaptation: Elbflorenz assigned 28 of 29 cases to grade 2 and Camma assigned 26 of 29 to grade 0. Within Task 2, collapse is concentrated at the smaller nodes, most clearly Elbflorenz at Center 2 with 8 of 9 predictions on a single grade, while Center 3, the largest node, is the least degenerate setting for all three submissions. Two distinct mechanisms are therefore at work. Failure to transfer under domain shift produces collapse in the global model itself, independent of adaptation, and unconstrained fine tuning on small, skewed local datasets produces collapse where local sample counts are lowest. Both inflate nominal F1 without reflecting genuine discrimination. As discussed in Section \ref{sec:datascale_discussion}, the EC metric is more robust to this artifact and should be considered the primary indicator of adaptation quality in imbalanced ordinal settings.

\subsection{Toward Personalized Federated Learning}
\label{sec:personalized_fl}
The results of both tasks point toward the same conclusion: a single globally aggregated model is a compromise that satisfies no center optimally. However, naive local fine-tuning, as evaluated in Task 2, does not resolve this, as unconstrained adaptation on skewed local data leads to catastrophic forgetting rather than genuine personalization, particularly at the smaller nodes. As shown in Section~\ref{sec:heterogeneity}, this is one of two routes to collapse: the global model itself may already fail to transfer to an unseen center, in which case no amount of constrained adaptation will recover discrimination. The recommendations below therefore address both, a globally federated component that resists center-specific drift, and a locally adapted component that cannot destabilize it. The path forward lies in structured personalized FL: approaches that separate a shared global representation from center-specific adaptation components, and train both with objectives that explicitly account for local data imbalance.

For Task 1 generalization, the FedSurg results support the use of a frozen pretrained temporal backbone as the globally federated component. By fixing backbone weights across federation rounds, center-specific visual variance cannot propagate into the shared representation, preserving its generalizability to unseen centers. Only lightweight task-specific heads need to be aggregated globally, substantially reducing both communication cost and the risk of conflicting updates between heterogeneous nodes.

For Task 2 adaptation, parameter-efficient fine-tuning (PEFT) methods offer a principled alternative to full local fine-tuning. Methods such as LoRA or adapter layers constrain the parameter space available for local updates, limiting the degree to which a small, imbalanced local dataset can destabilize the global representation, while still enabling meaningful center-specific adjustment. This approach is particularly relevant given the data regime of this challenge, where local node datasets ranged from 33 to 80 videos.

Looking further ahead, surgical video foundation models represent a promising direction for initializing the shared backbone \cite{yang_large-scale_2025, schmidgall_general_2024}. However, currently available surgical foundation models are trained on limited procedure-specific data, and general-purpose video foundation models produce feature spaces too broad for the fine-grained visual distinctions required in intraoperative grading tasks. Recent work combining FL with endoscopic foundation model pretraining \cite{kirchner_federated_2025} suggests a more principled path: federating the pretraining process itself to build privacy-preserving, domain-specific representations from the outset. The integration of such approaches with PEFT-based local adaptation remains a hypothesis for future evaluation rather than a validated recommendation.

The exploratory PEFT baselines included in this work provide a first and partly contrary empirical test for these recommendations. The EndoViT+LoRA variant, which applies LoRA to a frozen frame-level backbone, collapsed to near majority-class prediction (Task~1 F1: 9.24\%), consistent with the interpretation that parameter-efficient adaptation cannot compensate for the absence of temporal modeling in this data regime, though the two backbones also differ in pretraining domain (Section~\ref{sec:temporal}). Under the ordinal cost this variant is nevertheless the cheaper of the two on Task~1 (17.24\% against 22.76\%), which is what collapse onto a central grade produces, and neither improves on a constant predictor there. The ViViT+LoRA variant, applying LoRA to a frozen temporal backbone, achieved 15.93\% F1 on Task~1 and 29.57\% F1 at Center~3 in Task~2, approaching the centralized baseline (32.67\%) at the largest local node. Performance at smaller centers remained limited (Center~1: 12.22\%, Center~2: 22.62\%), consistent with the quantity-skew hypothesis: PEFT-based local adaptation is most effective where sufficient local samples are available to support stable fine-tuning.

\subsection{Implications for Task Design}
Beyond algorithmic choices, the FedSurg results suggest that the design of the clinical prediction task itself should inform the choice of FL strategy. This dimension has received limited attention in the literature. The consequences of prediction errors differ fundamentally between tasks, and these differences have direct implications for how generalization and adaptation should be balanced.

Consider two contrasting task types. In high-sensitivity settings, for example the detection of malignant lesions in colonoscopy where false negatives carry direct patient harm, global generalization to unseen centers is critical. FL strategies that prioritize robustness across non-IID distributions, such as flat-minima optimization approaches like FedSAM, are preferable over those that optimize local performance at the cost of cross-center generalizability. The fine-grained 6-class inflammation grading evaluated in this challenge approaches the opposite end of the spectrum: overcalling severity could influence clinical decision-making, placing greater weight on specificity. Here, the instability of local adaptation observed in Task 2 is particularly concerning. Classifier collapse, the reversion to a predictor that assigns nearly all cases to a single grade, is not only a performance failure but a specific risk in high specificity grading contexts. As shown in Section \ref{sec:heterogeneity}, collapse in this challenge arises from two sources: failure of the global model to transfer to an unseen center, and unconstrained fine tuning on small, imbalanced local datasets. A collapsed model may retain a superficially acceptable nominal score while providing no clinically actionable discrimination, and the direction of collapse determines the clinical consequence: collapse toward a low grade systematically undercalls severity, whereas collapse toward a high grade overcalls it. Neither failure mode is detectable from an aggregate F1-score alone.

These observations lead to a practical hypothesis for future work: FL strategy selection should be guided not only by dataset characteristics, but by an explicit analysis of the clinical error cost structure of the target task. High-sensitivity tasks call for globally robust aggregation; high-specificity tasks require adaptation strategies that prevent classifier collapse, such as the PEFT-based approaches discussed in Section~\ref{sec:personalized_fl}. Systematic evaluation of this hypothesis across diverse surgical AI tasks remains an important open problem.

\subsection{Limitations and Future Directions}
The limitations of this challenge should be read against the state of the field. A recent systematic review of 188 surgical scene understanding studies reports a median of 65 videos and 40 patients per study, with 70.7\% of studies relying on a single institution and 59.0\% on laparoscopic cholecystectomy~\cite{carstens_artificial_2025}. The field is also dominated by frame-level instrument and anatomy detection tasks, which are substantially easier than patient-level ordinal grading and on which higher scores are routinely reported. We consider the reporting of weak absolute performance on a clinically meaningful task, measured against chance and constant references, more useful to realistic expectations of clinical AI performance than stronger numbers on tasks of limited clinical relevance would be.

That said, the specific constraints of this edition are real and bear directly on what can be concluded from it. The evaluation rests on 153 training and 70 test videos across four centers, which constrains what it can resolve. Paired permutation tests over the test cases reach significance only for large differences, and no adaptation comparison reaches significance on either metric. Bootstrap confidence intervals on individual scores span roughly ten percentage points, and the Task~2 rank ordering is not reproducible under resampling. Reported point estimates should therefore be read as indicative rather than as reliable estimates of relative method quality, and the leaderboard is reported for completeness rather than as evidence of submission superiority. No clinically acceptable performance threshold has been established for intraoperative appendicitis grading, so the reported scores cannot be assessed against a target; the six-class ordinal label itself carries substantial ambiguity, with a weighted inter-annotator kappa of $0.614$ reported for the complete dataset \cite{kolbinger_appendix300_2025}, which bounds attainable performance independently of the learning paradigm.

Despite 24 registered teams, only three complete submissions were received, reflecting both the niche character of FL within surgical AI and the high technical barriers to participation. Several aspects of the challenge design likely contributed. The requirement to submit a containerized end-to-end federated pipeline raised the implementation burden substantially relative to a typical segmentation or detection challenge; expertise in federated optimization and in surgical video analysis rarely coincides within a single group; and the public and private data partition, introduced to discourage centralized development, prevented conventional development workflows. Future editions should consider enlarging the development subset or releasing a separate development split, since the guarantee against centralized training derives from the organizers executing the final federated run rather than from withholding data. Providing feedback runs on the complete training data ahead of the final deadline would allow participants to observe federated convergence and tune accordingly.

The three submissions differed in backbone architecture, frame sampling, loss function and aggregation strategy simultaneously. Observed performance differences therefore cannot be attributed to any single design choice, and in particular the contribution of the FL strategy cannot be separated from that of the architecture. The parameter-efficient baselines discussed in Section~\ref{sec:temporal} come closest to a controlled comparison here, though their backbones differ in pretraining corpus as well as in temporal modeling, and the organizer-provided baselines are post-hoc analyses rather than competitive entries.

The dataset, as a preliminary subset of Appendix300 \cite{kolbinger_appendix300_2025}, was insufficient in scale for robust federated training, and the simulated FL setup, where all data resided centrally during evaluation, does not capture the full complexity of real-world distributed deployments, including network latency, node availability, and hardware heterogeneity. A further practical lesson concerns the evaluation infrastructure itself. Frame extraction does not yield an identical frame count for every recording, since the available footage around the annotated timestamp is occasionally shorter than the nominal window. An evaluation harness that filters inputs on an expected frame count will silently discard such cases, which can produce divergent test sets between submission and baseline pipelines without any error being raised. Challenge harnesses should therefore assert explicitly on test set size and composition at evaluation time rather than skipping inputs that fail a format check. This occurred in the present challenge. One of the ten Center~1 test cases was initially skipped in the evaluation of the participant submissions, while the organizer baselines, which sample frames at a fixed rate from the full recording rather than from a fixed index range, were always evaluated on all ten. Predictions for that case were subsequently obtained from all three teams, and every result reported here is computed on the complete test set of 70 cases. The rank ordering is unaffected.

Future work should prioritize scaling the dataset across a broader network of institutions and transitioning from simulated to genuine multi-institutional FL. On the algorithmic side, imbalance-aware objectives, such as federated focal loss or client-weighted aggregation, and personalized aggregation methods that account for quantity skew are necessary next steps. Self-supervised pretraining on large-scale surgical video collections, ideally in a federated manner \cite{kirchner_federated_2025}, offers a promising route toward more transferable representations without compromising data privacy.

\section{Conclusion}

The FedSurg Challenge establishes the first benchmark for FL in surgical video classification, evaluated here as a proof-of-concept using a multi-center dataset of laparoscopic appendectomies. The results reveal that temporal modeling is the architectural factor most consistently associated with cross-center generalization, though the effect is not uniform across metrics in the controlled baseline comparison, while absolute performance remains far below clinical usability under both centralized and decentralized training, and several configurations fail to improve on constant or chance-level prediction. We do not establish a quantified cost of decentralization: at this sample size the centralized and federated baselines are not statistically separable from one another, and only large differences between methods are resolvable at all. Local fine-tuning without imbalance-aware objectives leads to classifier collapse rather than genuine personalization, as does failure of the global model to transfer to an unseen center, underscoring the need for structured personalized FL with parameter-efficient adaptation. By characterizing these failure modes, and the resolution limits of the evaluation itself, this work provides a methodological reference point for the design of robust, privacy-preserving AI systems for surgical video analysis and motivates future benchmarking on tasks with broader clinical scope.

\section*{Acknowledgments}
This work was co-funded by the European Union through NEARDATA under grant agreement ID 101092644, the German Research Foundation (DFG, Deutsche Forschungsgemeinschaft) as part of Germany’s Excellence Strategy – EXC 2050/1 – Project ID 390696704 – Cluster of Excellence “Centre for Tactile Internet with Human-in-the-Loop” (CeTI) of Dresden University of Technology, and the Federal Ministry of Education and Research of Germany in the program of “Souverän. Digital. Vernetzt.”, a joint project 6G-life with the project identification number 16KISK001K.

Furthermore, FRK receives support from the German Cancer Research Center (CoBot 2.0), the Joachim Herz Foundation (Add-On Fellowship for Interdisciplinary Life Science), the Central Indiana Corporate Partnership AnalytiXIN Initiative, the Evan and Sue Ann Werling Pancreatic Cancer Research Fund, and the Indiana Clinical and Translational Sciences Institute (EPAR4157) funded, in part, by Grant Number UM1TR004402 from the National Institutes of Health, National Center for Advancing Translational Sciences, Clinical and Translational Sciences Award. The content is solely the responsibility of the authors and does not necessarily represent the official views of the National Institutes of Health.

Team Camma (NP, JA) was supported by French State Funds managed by the Agence Nationale de la Recherche (ANR) under Grants ANR-22-FAI1-0001 (Project DAIOR) and ANR-10-IAHU-02 (IHU Strasbourg).

\section*{Conflict of Interest}
JNK declares consulting services for Bioptimus, France; Panakeia, UK; AstraZeneca, UK; and MultiplexDx, Slovakia. Furthermore, he holds shares in StratifAI, Germany, Synagen, Germany, Ignition Lab, Germany; has received an institutional research grant by GSK; and has received honoraria by AstraZeneca, Bayer, Daiichi Sankyo, Eisai, Janssen, Merck, MSD, BMS, Roche, Pfizer, and Fresenius. 

FRK declares advisory roles for Radical Healthcare, USA; and the Surgical Data Science Collective, USA, and has received research funding from Novartis. 

SS received speaker's fees from Stryker Corporation.

AR, NP, SS, DS, and SBa serve as an Associate Editor for the Medical Image Analysis journal.

NP is co-founder and owns shares in Scialytics SAS.

\section*{Declaration of generative AI and AI-assisted technologies in the writing process.}
During the preparation of this work, the author(s) used ChatGPT, Gemini 2.5 Pro, Claude Code, and DeepL Write in order to improve the readability and language of the manuscript. After using these tools/services, the author(s) reviewed and edited the content as needed and take full responsibility for the content of the published article.

\appendix
\section{ Dataset and Annotation Protocol}\label{A:appanno}
The challenge dataset is a preliminary subset of the publicly available Appendix300 dataset \cite{kolbinger_appendix300_2025}. A CSV file detailing the samples used in the challenge, along with two recordings that were excluded from the final dataset but used in the challenge, is deposited for reproducibility\footnote{Available at \href{https://doi.org/10.25532/OPARA-1534}{https://doi.org/10.25532/OPARA-1534}}. Both deposits are available under a CC BY license and require citation of the Appendix300 publication\footnote{Available at \href{https://doi.org/10.25532/OPARA-1173}{https://doi.org/10.25532/OPARA-1173}}. The annotation protocol is available in \cite{kolbinger_appendix300_2025}. Usage of the data for individual publications was prohibited before the release of this study and the Appendix300 dataset.

\section{Challenge Rules}\label{A:rules}
During registration, participants signed the EndoVis rules document\footnote{Available at \href{https://www.dropbox.com/scl/fi/5dm24ohyv65gi9d4d9dv0/EndoVis_Rules.pdf?rlkey=0rzddq688zc65fkhe5tcdtwlj&dl=0}{https://tinyurl.com/nhb5z6a3}}.

\section{Challenge Organization}\label{A:challenge_orga}
This work was jointly organized by four parties. Fiona R. Kolbinger, from the Department of Visceral, Thoracic and Vascular Surgery, Faculty of Medicine and University Hospital Carl Gustav Carus of the TUD Dresden University of Technology and the Weldon School of Biomedical Engineering, Purdue University coordinated the challenge data collection from four hospitals in Germany. Oliver L. Saldanha and Jakob N. Kather, also from the Else Kröner Fresenius Center for Digital Health, supported the data aggregation from the technical point of view and obtained the source data from these centers. Max Kirchner, Alexander C. Jenke, Sebastian Bodenstedt, and Stefanie Speidel, from the Department of Translational Surgical Oncology (TSO) of the National Center for Tumor Diseases (NCT) Dresden, supported data aggregation and carried out the challenge organization and the technical implementation, including preprocessing, data provision, participant registration and administration, submission handling, evaluation, and results presentation. The NearData Horizon Europe program provided €500 in prize money, equally distributed across both tasks. If a single team achieved the top score in both tasks, they were eligible to receive both awards.

\section{Submission Instructions} \label{A:submission_process}
The submission process was described in detail on the official challenge website\footnote{Available at \href{https://www.synapse.org/Synapse:syn53137385/wiki/625370}{https://www.synapse.org/Synapse:syn53137385/wiki/625370}}. To support development, an example setup based on the FL Flower framework \cite{beutel_flower_2022} was provided through a GitLab repository\footnote{Available at \href{https://gitlab.com/nct\_tso\_public/challenges/miccai2024/FedSurg24}{https://gitlab.com/nct\_tso\_public/challenges/miccai2024/FedSurg24}}. The evaluation framework was made transparent through another repository\footnote{Available at \href{https://gitlab.com/nct\_tso\_public/challenges/miccai2024/snippet}{https://gitlab.com/nct\_tso\_public/challenges/miccai2024/snippet}\label{fedsurg}}, which included the source code for metric computation as well as the ranking scripts (bootstrapping, statistical testing, and related plots).

Participants submitted their solutions as Docker containers via the challenge website. Docker Compose scripts with additional code were also accepted, provided they encapsulated the complete FL algorithm for both training and inference. Submissions had to run fully automatically without user interaction. Participants received email notifications about submission status and were allowed unlimited resubmissions until the final deadline. While the organizers do not distribute Docker images, teams were encouraged to release their code publicly.

\section{Challenge design document} \label{A:design_doc}
See Supplementary file S1.

\section{Ethics approval}\label{A:ethics}
This study was prospectively reviewed and approved by the Institutional Review Board of the TUD Dresden University of Technology, Germany (approval number: BO-EK-332072022, approval date: August 4, 2022). The corresponding study was prospectively registered at the German Registry of Clinical Trials (DRKS, \href{https://drks.de/search/de/trial/DRKS00030874}{URL}, registration ID DRKS00030874).

\section{Additional Bootstrapping and Resampling Stability Results}
\label{A:bootstrapping}

The third column group of each table reports a bootstrap discordance value, computed by applying the Wilcoxon signed-rank test to the two methods' metric values across the 10{,}000 bootstrap replicates. Because the test is applied to resampling replicates rather than to the test cases, its sample size is set by the number of replicates and not by the data, and the resulting values become arbitrarily small for any consistent difference as the number of replicates increases. A small value indicates that one method exceeds the other in nearly every replicate, and a value reported as $0.000$ indicates that this holds so consistently that the statistic underflows double precision ($<10^{-308}$); intermediate values such as $8.38 \times 10^{-90}$ indicate a slightly less consistent ordering, but the difference between such magnitudes carries no interpretable meaning. These values therefore describe how consistently one method exceeds another under resampling of a fixed test set, and are reported alongside the rank frequencies and win probabilities as stability diagnostics rather than as hypothesis tests. Inference over the actual test cases is provided by the paired permutation tests in Table~\ref{tab:permutation}.

\begin{table}[h!]
\centering
\scriptsize
\setlength{\tabcolsep}{3pt}
\renewcommand{\arraystretch}{1.05}
\caption{Bootstrap rank frequency, win probability, and bootstrap discordance
(descriptive) for both tasks (F1-score $\uparrow$). Values in the discordance
group are computed over 10{,}000 bootstrap replicates and describe resampling
stability. They are not population-level $p$-values; see
Section~\ref{sec:ranking} and the permutation tests in
Table~\ref{tab:permutation}.}
\label{tab:bootstrap_teams_f1}
\begin{tabular}{cc|l|ccc|ccc|ccc}
\hline
Task & Center & Team & \multicolumn{3}{c|}{Rank Freq.} & \multicolumn{3}{c|}{Win Prob.} & \multicolumn{3}{c}{\shortstack{Bootstrap discordance\\(descriptive)}} \\
 & & & 1 & 2 & 3 & Cam & Elb & San & Cam & Elb & San \\
\hline
1 & 4 & Cam   & 0.0000 & 0.2491 & \textbf{0.7509} & NaN & 0.2481 & 0.0000 & NaN & 0.0000 & 0.0000 \\
1 & 4 & Elb   & 0.0077 &\textbf{ 0.7442} & 0.2481 & 0.7509 & NaN & 0.0077 & 0.0000 & NaN & 0.0000 \\
1 & 4 & San   & \textbf{0.9923} & 0.0077 & 0.0000 & 1.0000 & 0.9923 & NaN & 0.0000 & 0.0000 & NaN \\
\hline
2 & Avg & Cam   & \textbf{0.7852} & 0.1429 & 0.0719 & NaN & 0.8706 & 0.8427 & NaN & 0.0000 & 0.0000 \\
2 & Avg & Elb   & 0.0909 & 0.3710 & \textbf{0.5381} & 0.1294 & NaN & 0.4233 & 0.0000 & NaN & $1.31\times 10^{-70}$ \\
2 & Avg & San   & 0.1239 & \textbf{0.4862} & 0.3899 & 0.1573 & 0.5766 & NaN & 0.0000 & $1.31\times 10^{-70}$ & NaN \\
\hline
\end{tabular}
\end{table}

\begin{table}[h!]
\centering
\scriptsize
\setlength{\tabcolsep}{3pt}
\renewcommand{\arraystretch}{1.05}
\caption{Bootstrap rank frequency, win probability, and bootstrap discordance (descriptive) for both tasks (EC $\downarrow$). Values in the discordance group are computed over 10{,}000 bootstrap replicates and describe resampling stability. They are not population-level $p$-values; see Section~\ref{sec:ranking} and the permutation tests in Table~\ref{tab:permutation}. At Task~2 both Santhi and Elbflorenz take rank~1 most frequently, in $40.4\%$ and $36.0\%$ of replicates respectively, so the modal ranking does not reproduce the nominal ordering.}
\label{tab:bootstrap_teams_ec}
\begin{tabular}{cc|l|ccc|ccc|ccc}
\hline
Task & Center & Team & \multicolumn{3}{c|}{Rank Freq.} & \multicolumn{3}{c|}{Win Prob.} & \multicolumn{3}{c}{\shortstack{Bootstrap discordance\\(descriptive)}} \\
 & & & 1 & 2 & 3 & Cam & Elb & San & Cam & Elb & San \\
\hline
1 & 4 & Cam   & 0.0000 & 0.0000 & \textbf{1.0000} & NaN & 0.0000 & 0.0000 & NaN & 0.0000 & 0.0000 \\
1 & 4 & Elb   & 0.0027 & \textbf{0.9973} & 0.0000 & 1.0000 & NaN & 0.0021 & 0.0000 & NaN & 0.0000 \\
1 & 4 & San   & \textbf{0.9979} & 0.0021 & 0.0000 & 1.0000 & 0.9973 & NaN & 0.0000 & 0.0000 & NaN \\
\hline
2 & Avg & Cam   & 0.2368 & 0.3142 & \textbf{0.4490} & NaN & 0.4227 & 0.3640 & NaN & $8.38\times 10^{-90}$ & $4.68\times 10^{-226}$ \\
2 & Avg & Elb   & \textbf{0.3600} & 0.3230 & 0.3170 & 0.5772 & NaN & 0.4652 & $8.38\times 10^{-90}$ & NaN & $4.80\times 10^{-20}$ \\
2 & Avg & San   & \textbf{0.4036} & 0.3636 & 0.2328 & 0.6350 & 0.5343 & NaN & $4.68\times 10^{-226}$ & $4.80\times 10^{-20}$ & NaN \\
\hline
\end{tabular}
\end{table}

\section{Further Analysis Results with Baselines}

\begin{table}[h!]
\centering
\tiny
\setlength{\tabcolsep}{3pt}
\renewcommand{\arraystretch}{1.05}
\caption{Comprehensive ranking summary for Task 1, Task 2, and overall performance including contextual baselines. Lower rank indicates better performance. The two parameter-efficient baselines are not ranked here; as stated in Section~\ref{sec:baselines} they are reported as point estimates only.}
\label{tab:appendix_combined_ranking}
\begin{tabular}{l|ccc|ccc|c|c}
\hline
\textbf{Team} & \multicolumn{3}{c}{\textbf{Task 1}} & \multicolumn{3}{c}{\textbf{Task 2}} & \textbf{Overall} & \textbf{Final} \\
& \textbf{EC} & \textbf{F1} & \textbf{Avg} & \textbf{EC} & \textbf{F1} & \textbf{Avg} & \textbf{Avg} & \textbf{Rank} \\ \hline
Camma      & 5 & 5 & 5 & 5 & 2 & 3 & 4.00 & 3 \\
Elbflorenz & 4 & 4 & 4 & 4 & 5 & 4 & 4.00 & 3 \\
Santhi     & 2 & 2 & 2 & 3 & 4 & 3 & 2.50 & 2 \\
Cen        & 1 & 1 & 1 & 1 & 1 & 1 & 1.00 & 1 \\
SWARM      & 3 & 3 & 3 & 2 & 3 & 2 & 2.50 & 2 \\ \hline
\end{tabular}
\end{table}

\subsection{Detailed Bootstrap Analysis Tables}

\paragraph{Task 1: Generalization (F1-score)}
Tables~\ref{tab:bootstrap_task1_f1_ranks}--\ref{tab:bootstrap_task1_f1_wilcoxon} report rank frequency, pairwise win probabilities, and bootstrap discordance for Task~1 F1-score. Rankings are comparatively stable: the centralized baseline holds rank~1 in 65.7\% of bootstrap iterations and Santhi in 32.1\%, so those two are not resolved by the resampling. The paired permutation test over the 29 Center-4 cases agrees ($p = 0.671$ for Santhi vs.\ Cen). It does separate the two lowest-placed submissions from the leaders: Camma and Elbflorenz each differ significantly from Santhi and from Cen ($p \le 0.047$), and Camma additionally from Swarm ($p = 0.005$), though Elbflorenz does not ($p = 0.122$). Camma and Elbflorenz are indistinguishable from each other ($p = 0.592$).
\begin{table}[h!]
\centering
	\tiny
\caption{Task 1 F1-score $\uparrow$: Rank frequency distribution.}
\label{tab:bootstrap_task1_f1_ranks}
\begin{tabular}{lccccc}
\hline
	\textbf{Team} & \textbf{Rank 1} & \textbf{Rank 2} & \textbf{Rank 3} & \textbf{Rank 4} & \textbf{Rank 5} \\
\hline
Camma       & 0.0000 & 0.0000 & 0.0001 & 0.2490 & 0.7509 \\
Elbflorenz  & 0.0001 & 0.0023 & 0.0126 & 0.7369 & 0.2481 \\
Santhi      & 0.3206 & 0.4332 & 0.2397 & 0.0065 & 0.0000 \\
Cen         & 0.6566 & 0.2953 & 0.0480 & 0.0001 & 0.0000 \\
SWARM       & 0.0242 & 0.2689 & 0.6984 & 0.0085 & 0.0000 \\
\hline
\end{tabular}
\end{table}

\begin{table}[h!]
\centering
	\tiny
\caption{Task 1 F1-score $\uparrow$: Pairwise win probabilities $P(\text{team}_i > \text{team}_j)$.}
\label{tab:bootstrap_task1_f1_pairwise}
\begin{tabular}{lccccc}
\hline
	\textbf{Team} & \textbf{Camma} & \textbf{Elbflorenz} & \textbf{Santhi} & \textbf{Cen} & \textbf{SWARM} \\
\hline
Camma       & --- & 0.2481 & 0.0000 & 0.0000 & 0.0001 \\
Elbflorenz  & 0.7509 & --- & 0.0077 & 0.0003 & 0.0095 \\
Santhi      & 1.0000 & 0.9923 & --- & 0.3266 & 0.7477 \\
Cen         & 1.0000 & 0.9997 & 0.6728 & --- & 0.9339 \\
SWARM       & 0.9999 & 0.9905 & 0.2516 & 0.0647 & --- \\
\hline
\end{tabular}
\end{table}

\begin{table}[h!]
\centering
	\tiny
\caption{Task 1 F1-score $\uparrow$: Bootstrap discordance (descriptive). Values are computed over 10{,}000 bootstrap replicates and describe resampling stability. They are not population-level $p$-values; see Section~\ref{sec:ranking} and the permutation tests in Table~\ref{tab:permutation}. Every entry underflows double precision ($<10^{-308}$), including pairs the permutation test cannot separate.}
\label{tab:bootstrap_task1_f1_wilcoxon}
\begin{tabular}{lccccc}
\hline
	\textbf{Team} & \textbf{Camma} & \textbf{Elbflorenz} & \textbf{Santhi} & \textbf{Cen} & \textbf{SWARM} \\
\hline
Camma       & --- & 0.000 & 0.000 & 0.000 & 0.000 \\
Elbflorenz  & 0.000 & --- & 0.000 & 0.000 & 0.000 \\
Santhi      & 0.000 & 0.000 & --- & 0.000 & 0.000 \\
Cen         & 0.000 & 0.000 & 0.000 & --- & 0.000 \\
SWARM       & 0.000 & 0.000 & 0.000 & 0.000 & --- \\
\hline
\end{tabular}
\end{table}

\paragraph{Task 1: Generalization (Expected Cost)}
Tables~\ref{tab:bootstrap_task1_ec_ranks}--\ref{tab:bootstrap_task1_ec_wilcoxon} report the equivalent analysis for the EC metric, which gives the sharpest separation anywhere in the challenge. Camma retains the last rank in 100\% of iterations and Elbflorenz the fourth in 98.6\%, and the permutation test confirms this: Camma differs from all four other methods ($p = 10^{-4}$) and Elbflorenz from Santhi, Cen and SWARM ($p = 0.013$, $0.004$ and $0.039$). The leading three are not separated from one another --- Santhi, Cen and SWARM exchange the top three positions across replicates, and no permutation comparison among them approaches significance (Santhi vs.\ Cen $p = 0.638$, Santhi vs.\ SWARM $p = 1.000$, Cen vs.\ SWARM $p = 0.357$).
\begin{table}[h!]
\centering
	\tiny
\caption{Task 1 EC $\downarrow$: Rank frequency distribution. Santhi's modal rank (3) differs from its nominal rank (2).}
\label{tab:bootstrap_task1_ec_ranks}
\begin{tabular}{lccccc}
\hline
	\textbf{Team} & \textbf{Rank 1} & \textbf{Rank 2} & \textbf{Rank 3} & \textbf{Rank 4} & \textbf{Rank 5} \\
\hline
Camma       & 0.0000 & 0.0000 & 0.0000 & 0.0000 & 1.0000 \\
Elbflorenz  & 0.0002 & 0.0011 & 0.0124 & 0.9863 & 0.0000 \\
Santhi      & 0.2619 & 0.3560 & 0.3810 & 0.0011 & 0.0000 \\
Cen         & 0.7452 & 0.2309 & 0.0238 & 0.0001 & 0.0000 \\
SWARM       & 0.0890 & 0.3882 & 0.5132 & 0.0096 & 0.0000 \\
\hline
\end{tabular}
\end{table}

\begin{table}[h!]
\centering
	\tiny
\caption{Task 1 EC $\downarrow$: Pairwise win probabilities.}
\label{tab:bootstrap_task1_ec_pairwise}
\begin{tabular}{lccccc}
\hline
	\textbf{Team} & \textbf{Camma} & \textbf{Elbflorenz} & \textbf{Santhi} & \textbf{Cen} & \textbf{SWARM} \\
\hline
Camma       & --- & 0.0000 & 0.0000 & 0.0000 & 0.0000 \\
Elbflorenz  & 1.0000 & --- & 0.0021 & 0.0002 & 0.0097 \\
Santhi      & 1.0000 & 0.9973 & --- & 0.2094 & 0.5482 \\
Cen         & 1.0000 & 0.9996 & 0.7287 & --- & 0.8855 \\
SWARM       & 1.0000 & 0.9879 & 0.3905 & 0.0692 & --- \\
\hline
\end{tabular}
\end{table}

\begin{table}[h!]
\centering
	\tiny
\caption{Task 1 EC $\downarrow$: Bootstrap discordance (descriptive). Values are computed over 10{,}000 bootstrap replicates and describe resampling stability. They are not population-level $p$-values; see Section~\ref{sec:ranking} and the permutation tests in Table~\ref{tab:permutation}. Santhi and SWARM receive $1.40 \times 10^{-95}$ here, while the permutation test over the 29 test cases returns $p = 1.000$ for that pair.}
\label{tab:bootstrap_task1_ec_wilcoxon}
\begin{tabular}{lccccc}
\hline
	\textbf{Team} & \textbf{Camma} & \textbf{Elbflorenz} & \textbf{Santhi} & \textbf{Cen} & \textbf{SWARM} \\
\hline
Camma       & --- & 0.000 & 0.000 & 0.000 & 0.000 \\
Elbflorenz  & 0.000 & --- & 0.000 & 0.000 & 0.000 \\
Santhi      & 0.000 & 0.000 & --- & 0.000 & $1.40 \times 10^{-95}$ \\
Cen         & 0.000 & 0.000 & 0.000 & --- & 0.000 \\
SWARM       & 0.000 & 0.000 & $1.40 \times 10^{-95}$ & 0.000 & --- \\
\hline
\end{tabular}
\end{table}

\paragraph{Task 2: Adaptation (F1-score)}
Tables~\ref{tab:bootstrap_task2_f1_ranks}--\ref{tab:bootstrap_task2_f1_wilcoxon} show markedly higher rank volatility compared to Task~1. No method retains its nominal rank in more than 72.3\% of iterations (Cen), and among the three participant submissions none exceeds 52.0\% (Elbflorenz); Camma, the highest-placed submission on this metric, holds its nominal rank~2 in only 36.7\% of iterations and takes rank~1 in 17.8\%. The pairwise win probabilities between Camma, Santhi, and SWARM indicate substantial overlap, reflecting the instability of local adaptation across centers. No pairwise permutation comparison on Task~2 F1 reaches significance; the smallest is $p = 0.073$ for Santhi vs.\ Cen.
\begin{table}[htbp]
\centering
	\tiny
\caption{Task 2 F1-score $\uparrow$: Rank frequency distribution.}
\label{tab:bootstrap_task2_f1_ranks}
\begin{tabular}{lccccc}
\hline
	\textbf{Team} & \textbf{Rank 1} & \textbf{Rank 2} & \textbf{Rank 3} & \textbf{Rank 4} & \textbf{Rank 5} \\
\hline
Camma       & 0.1776 & 0.3672 & 0.2812 & 0.1083 & 0.0657 \\
Elbflorenz  & 0.0063 & 0.0395 & 0.0991 & 0.3351 & 0.5200 \\
Santhi      & 0.0048 & 0.0593 & 0.1353 & 0.4257 & 0.3749 \\
Cen         & 0.7232 & 0.1820 & 0.0792 & 0.0125 & 0.0031 \\
SWARM       & 0.0881 & 0.3521 & 0.4051 & 0.1185 & 0.0362 \\
\hline
\end{tabular}
\end{table}

\begin{table}[htbp]
\centering
	\tiny
\caption{Task 2 F1-score $\uparrow$: Pairwise win probabilities.}
\label{tab:bootstrap_task2_f1_pairwise}
\begin{tabular}{lccccc}
\hline
	\textbf{Team} & \textbf{Camma} & \textbf{Elbflorenz} & \textbf{Santhi} & \textbf{Cen} & \textbf{SWARM} \\
\hline
Camma       & --- & 0.8706 & 0.8427 & 0.2094 & 0.5600 \\
Elbflorenz  & 0.1294 & --- & 0.4233 & 0.0183 & 0.1059 \\
Santhi      & 0.1573 & 0.5766 & --- & 0.0152 & 0.1441 \\
Cen         & 0.7906 & 0.9817 & 0.9848 & --- & 0.8526 \\
SWARM       & 0.4400 & 0.8941 & 0.8558 & 0.1474 & --- \\
\hline
\end{tabular}
\end{table}

\begin{table}[htbp]
\centering
	\tiny
\caption{Task 2 F1-score $\uparrow$: Bootstrap discordance (descriptive). Values are computed over 10{,}000 bootstrap replicates and describe resampling stability. They are not population-level $p$-values; see Section~\ref{sec:ranking} and the permutation tests in Table~\ref{tab:permutation}. No Task~2 F1 comparison reaches $p < 0.05$ under the permutation test.}
\label{tab:bootstrap_task2_f1_wilcoxon}
\begin{tabular}{lccccc}
\hline
	\textbf{Team} & \textbf{Camma} & \textbf{Elbflorenz} & \textbf{Santhi} & \textbf{Cen} & \textbf{SWARM} \\
\hline
Camma       & --- & 0.000 & 0.000 & 0.000 & $3.71 \times 10^{-42}$ \\
Elbflorenz  & 0.000 & --- & $1.31 \times 10^{-70}$ & 0.000 & 0.000 \\
Santhi      & 0.000 & $1.31 \times 10^{-70}$ & --- & 0.000 & 0.000 \\
Cen         & 0.000 & 0.000 & 0.000 & --- & 0.000 \\
SWARM       & $3.71 \times 10^{-42}$ & 0.000 & 0.000 & 0.000 & --- \\
\hline
\end{tabular}
\end{table}

\paragraph{Task 2: Adaptation (Expected Cost)}
Tables~\ref{tab:bootstrap_task2_ec_ranks}--\ref{tab:bootstrap_task2_ec_wilcoxon} report Task~2 EC bootstrap results, where rank instability is most pronounced. No method retains its nominal rank in even 40\% of iterations: the best-retained is Cen's at 37.9\%, followed by Camma's at 36.5\%, with the remaining three below 26\%. Two further observations underline the instability: Camma takes rank~1 in 8.6\% of iterations despite placing last on the point estimate, and Elbflorenz's modal rank (5) differs from its nominal rank (4). Win probabilities sit close to $0.5$ throughout, the most extreme being $0.756$ for Cen over Camma, and no permutation comparison approaches significance (smallest $p = 0.507$, Camma vs.\ Cen). In particular Santhi and SWARM differ by $0.62$ percentage points with a permutation $p$ of $0.844$ and heavily overlapping bootstrap intervals. EC-based adaptation rankings are not reproducible under resampling and carry essentially no information about submission superiority.
\begin{table}[htbp]
\centering
    \tiny
\caption{Task 2 EC $\downarrow$: Rank frequency distribution. Elbflorenz's modal rank (5) differs from its nominal rank (4).}
\label{tab:bootstrap_task2_ec_ranks}
\begin{tabular}{lccccc}
\hline
	\textbf{Team} & \textbf{Rank 1} & \textbf{Rank 2} & \textbf{Rank 3} & \textbf{Rank 4} & \textbf{Rank 5} \\
\hline
Camma       & 0.0861 & 0.1395 & 0.1716 & 0.2378 & 0.3650 \\
Elbflorenz  & 0.1535 & 0.1737 & 0.1965 & 0.2276 & 0.2487 \\
Santhi      & 0.1416 & 0.2055 & 0.2512 & 0.2467 & 0.1550 \\
Cen         & 0.3794 & 0.2267 & 0.1714 & 0.1283 & 0.0942 \\
SWARM       & 0.2414 & 0.2540 & 0.2090 & 0.1602 & 0.1354 \\
\hline
\end{tabular}
\end{table}

\begin{table}[htbp]
\centering
	\tiny
\caption{Task 2 EC $\downarrow$: Pairwise win probabilities.}
\label{tab:bootstrap_task2_ec_pairwise}
\begin{tabular}{lccccc}
\hline
	\textbf{Team} & \textbf{Camma} & \textbf{Elbflorenz} & \textbf{Santhi} & \textbf{Cen} & \textbf{SWARM} \\
\hline
Camma       & --- & 0.4227 & 0.3640 & 0.2440 & 0.3114 \\
Elbflorenz  & 0.5772 & --- & 0.4652 & 0.3275 & 0.3843 \\
Santhi      & 0.6350 & 0.5343 & --- & 0.3513 & 0.4074 \\
Cen         & 0.7557 & 0.6721 & 0.6473 & --- & 0.5911 \\
SWARM       & 0.6882 & 0.6152 & 0.5915 & 0.4084 & --- \\
\hline
\end{tabular}
\end{table}

\begin{table}[htbp]
\centering
	\tiny
\setlength{\tabcolsep}{3pt}
\caption{Task 2 EC $\downarrow$: Bootstrap discordance (descriptive). Values are computed over 10{,}000 bootstrap replicates and describe resampling stability. They are not population-level $p$-values; see Section~\ref{sec:ranking} and the permutation tests in Table~\ref{tab:permutation}. Every entry here is numerically extreme, yet no Task~2 EC comparison reaches $p < 0.05$ under the permutation test (smallest $p = 0.507$).}
\label{tab:bootstrap_task2_ec_wilcoxon}
\begin{tabular}{lccccc}
\hline
	\textbf{Team} & \textbf{Camma} & \textbf{Elbflorenz} & \textbf{Santhi} & \textbf{Cen} & \textbf{SWARM} \\
\hline
Camma       & --- & $8.38 \times 10^{-90}$ & $4.68 \times 10^{-226}$ & 0.000 & 0.000 \\
Elbflorenz  & $8.38 \times 10^{-90}$ & --- & $4.80 \times 10^{-20}$ & 0.000 & $5.23 \times 10^{-164}$ \\
Santhi      & $4.68 \times 10^{-226}$ & $4.80 \times 10^{-20}$ & --- & $2.56 \times 10^{-272}$ & $9.30 \times 10^{-95}$ \\
Cen         & 0.000 & 0.000 & $2.56 \times 10^{-272}$ & --- & $1.26 \times 10^{-89}$\\
SWARM       & 0.000 & $5.23 \times 10^{-164}$ & $9.30 \times 10^{-95}$ & $1.26 \times 10^{-89}$ & --- \\
\hline
\end{tabular}
\end{table}

\subsection{Paired Permutation Tests}

Table~\ref{tab:permutation} reports the inferential counterpart to the bootstrap diagnostics above. Twelve of the forty pairwise comparisons reach $p < 0.05$ and all twelve are on Task~1; no Task~2 comparison is significant on either metric. The comparisons among Santhi, Cen and SWARM are not significant on any task or metric, so the top of the ranking is not resolved by this evidence.

\begin{table}[htbp]
\centering
\tiny
\setlength{\tabcolsep}{4pt}
\renewcommand{\arraystretch}{1.05}
\caption{Paired permutation tests over the actual test cases. The exchangeable unit is the individual case: for each permutation the two methods' predictions are exchanged independently per case, with 10{,}000 permutations and a fixed seed. For Task~2 the exchange is performed within each center, the metric is recomputed per center, and the per-center values are then averaged with equal weight across centers. Sample sizes are $n = 29$ cases for Task~1 (Center~4) and $n = 41$ cases for Task~2 (10, 9 and 22 cases at Centers 1, 2 and 3). Observed differences are method~$A$ minus method~$B$ in percentage points; for F1 $\uparrow$ a positive value favours $A$, for EC $\downarrow$ a negative value favours $A$. Rows marked $\dagger$ also constitute the participants-only set of Tables~\ref{tab:bootstrap_teams_f1} and \ref{tab:bootstrap_teams_ec}; a pair's permutation result does not depend on which other methods are included. Significant comparisons ($p < 0.05$) in bold.}
\label{tab:permutation}
\begin{tabular}{ll|l|rr}
\hline
\textbf{Task} & \textbf{Metric} & \textbf{Comparison ($A$ vs.\ $B$)} & \textbf{Obs.\ diff.\ (pp)} & \textbf{Permutation $p$} \\
\hline
1 & F1 $\uparrow$ & Camma vs.\ Elbflorenz$^\dagger$ & $-3.07$  & 0.5919 \\
1 & F1 $\uparrow$ & Camma vs.\ Santhi$^\dagger$     & $-18.27$ & \textbf{0.0036} \\
1 & F1 $\uparrow$ & Camma vs.\ Cen                  & $-21.54$ & \textbf{0.0005} \\
1 & F1 $\uparrow$ & Camma vs.\ SWARM                & $-14.24$ & \textbf{0.0052} \\
1 & F1 $\uparrow$ & Elbflorenz vs.\ Santhi$^\dagger$& $-15.20$ & \textbf{0.0465} \\
1 & F1 $\uparrow$ & Elbflorenz vs.\ Cen             & $-18.48$ & \textbf{0.0084} \\
1 & F1 $\uparrow$ & Elbflorenz vs.\ SWARM           & $-11.17$ & 0.1221 \\
1 & F1 $\uparrow$ & Santhi vs.\ Cen                 & $-3.27$  & 0.6709 \\
1 & F1 $\uparrow$ & Santhi vs.\ SWARM               & $+4.03$  & 0.5876 \\
1 & F1 $\uparrow$ & Cen vs.\ SWARM                  & $+7.31$  & 0.2429 \\
\hline
1 & EC $\downarrow$ & Camma vs.\ Elbflorenz$^\dagger$ & $+33.10$ & \textbf{0.0001} \\
1 & EC $\downarrow$ & Camma vs.\ Santhi$^\dagger$     & $+44.83$ & \textbf{0.0001} \\
1 & EC $\downarrow$ & Camma vs.\ Cen                  & $+46.90$ & \textbf{0.0001} \\
1 & EC $\downarrow$ & Camma vs.\ SWARM                & $+44.14$ & \textbf{0.0001} \\
1 & EC $\downarrow$ & Elbflorenz vs.\ Santhi$^\dagger$& $+11.72$ & \textbf{0.0131} \\
1 & EC $\downarrow$ & Elbflorenz vs.\ Cen             & $+13.79$ & \textbf{0.0035} \\
1 & EC $\downarrow$ & Elbflorenz vs.\ SWARM           & $+11.03$ & \textbf{0.0390} \\
1 & EC $\downarrow$ & Santhi vs.\ Cen                 & $+2.07$  & 0.6380 \\
1 & EC $\downarrow$ & Santhi vs.\ SWARM               & $-0.69$  & 1.0000 \\
1 & EC $\downarrow$ & Cen vs.\ SWARM                  & $-2.76$  & 0.3566 \\
\hline
2 & F1 $\uparrow$ & Camma vs.\ Elbflorenz$^\dagger$ & $+6.58$  & 0.4208 \\
2 & F1 $\uparrow$ & Camma vs.\ Santhi$^\dagger$     & $+5.39$  & 0.4200 \\
2 & F1 $\uparrow$ & Camma vs.\ Cen                  & $-4.72$  & 0.4256 \\
2 & F1 $\uparrow$ & Camma vs.\ SWARM                & $+1.02$  & 0.8444 \\
2 & F1 $\uparrow$ & Elbflorenz vs.\ Santhi$^\dagger$& $-1.18$  & 0.8218 \\
2 & F1 $\uparrow$ & Elbflorenz vs.\ Cen             & $-11.30$ & 0.1161 \\
2 & F1 $\uparrow$ & Elbflorenz vs.\ SWARM           & $-5.56$  & 0.4024 \\
2 & F1 $\uparrow$ & Santhi vs.\ Cen                 & $-10.11$ & 0.0733 \\
2 & F1 $\uparrow$ & Santhi vs.\ SWARM               & $-4.37$  & 0.4060 \\
2 & F1 $\uparrow$ & Cen vs.\ SWARM                  & $+5.74$  & 0.3124 \\
\hline
2 & EC $\downarrow$ & Camma vs.\ Elbflorenz$^\dagger$ & $+0.81$ & 0.8355 \\
2 & EC $\downarrow$ & Camma vs.\ Santhi$^\dagger$     & $+1.20$ & 0.7431 \\
2 & EC $\downarrow$ & Camma vs.\ Cen                  & $+2.41$ & 0.5068 \\
2 & EC $\downarrow$ & Camma vs.\ SWARM                & $+1.82$ & 0.6513 \\
2 & EC $\downarrow$ & Elbflorenz vs.\ Santhi$^\dagger$& $+0.39$ & 0.9069 \\
2 & EC $\downarrow$ & Elbflorenz vs.\ Cen             & $+1.60$ & 0.6732 \\
2 & EC $\downarrow$ & Elbflorenz vs.\ SWARM           & $+1.01$ & 0.7845 \\
2 & EC $\downarrow$ & Santhi vs.\ Cen                 & $+1.21$ & 0.7313 \\
2 & EC $\downarrow$ & Santhi vs.\ SWARM               & $+0.62$ & 0.8444 \\
2 & EC $\downarrow$ & Cen vs.\ SWARM                  & $-0.59$ & 0.8701 \\
\hline
\end{tabular}
\end{table}

\subsection{Reference Classifiers}

Table~\ref{tab:reference_classifiers} gives the reference levels against which the reported scores should be read. All four references are constructed from training labels only and none uses test prevalence. \textsc{TrivialGlobal} predicts the pooled training mode (grade 3) everywhere; \textsc{TrivialLocal} predicts each center's own training mode (grade 2 at Center~1, grade 4 at Center~2, grade 3 at Center~3). \textsc{TrivialLocal} is undefined for Task~1, whose test center is held out and contributes no training labels of its own. \textsc{RandomUniform} draws each of the six grades with probability $1/6$; \textsc{RandomPrior} samples the training prior (pooled for Task~1, the center's own for Task~2). The two constant references are evaluated with 10{,}000 paired bootstrap replicates; the two stochastic references are simulated over 10{,}000 draws with a fresh prediction vector per draw, so they are reported with percentile intervals and are never passed through the ranking or win-probability machinery.

These references are floors rather than competitive levels, and several reported results fall below them. At Center~1 the constant \textsc{TrivialLocal} classifier attains an expected cost of 20.00\%, better than five of the seven evaluated methods; only Elbflorenz (18.00\%) and the EndoViT+LoRA baseline (17.78\%) improve on it. At Center~3 no participant submission reaches the 18.18\% expected cost of the constant classifier. Table~\ref{tab:reference_comparisons} lists the twenty comparisons in which a method falls below a chance classifier, covering eleven distinct method, test set and metric combinations.

\begin{table}[htbp]
\centering
\scriptsize
\setlength{\tabcolsep}{4pt}
\renewcommand{\arraystretch}{1.1}
\caption{Reference classifier levels, in percent. F1 $\uparrow$, EC $\downarrow$.
Constant references are point estimates; stochastic references are simulation
means with 95\% percentile intervals in brackets. Task~2 average is the
equal-weight mean over Centers 1--3. At Center~2 the edge position of grade~4
inflates the linear cost of TrivialLocal, so comparisons there should be made
against TrivialGlobal.}
\label{tab:reference_classifiers}

\textit{Constant references} (point estimates)\\[2pt]
\begin{tabular}{l ccc ccc}
\toprule
& \multicolumn{3}{c}{\textbf{TrivialGlobal}} & \multicolumn{3}{c}{\textbf{TrivialLocal}} \\
\cmidrule(lr){2-4}\cmidrule(lr){5-7}
\textbf{Test set} & Grade & F1 $\uparrow$ & EC $\downarrow$
                  & Grade & F1 $\uparrow$ & EC $\downarrow$ \\
\midrule
Task 1 Center 4 & 3 & 9.17 & 13.79 & n/a & n/a & n/a \\
\addlinespace[2pt]
Task 2 Center 1 & 3 & 3.03 & 28.00 & 2 & 9.52 & 20.00 \\
Task 2 Center 2 & 3 & 6.06 & 24.44 & 4 & 8.33 & 31.11 \\
Task 2 Center 3 & 3 & 9.68 & 18.18 & 3$^{*}$ & 9.68 & 18.18 \\
Task 2 average  & --- & 6.26 & 23.54 & --- & 9.18 & 23.10 \\
\bottomrule
\end{tabular}

\vspace{8pt}
\textit{Stochastic references} (simulation mean [95\% percentile interval])\\[2pt]
\begin{tabular}{l cc cc}
\toprule
& \multicolumn{2}{c}{\textbf{RandomUniform}} & \multicolumn{2}{c}{\textbf{RandomPrior}} \\
\cmidrule(lr){2-3}\cmidrule(lr){4-5}
\textbf{Test set} & F1 $\uparrow$ & EC $\downarrow$ & F1 $\uparrow$ & EC $\downarrow$ \\
\midrule
Task 1 Center 4 & 11.91 [2.4--23.9] & 33.55 [25.5--42.1] & 14.30 [5.5--26.1] & 25.52 [18.6--33.1] \\
\addlinespace[2pt]
Task 2 Center 1 & 12.29 [0.0--34.7] & 36.04 [20.0--52.0] & 14.78 [0.0--36.9] & 29.11 [16.0--42.0] \\
Task 2 Center 2 & 12.47 [0.0--36.1] & 35.84 [20.0--53.3] & 13.80 [0.0--34.4] & 29.80 [15.6--44.4] \\
Task 2 Center 3 & 13.83 [2.4--30.4] & 34.20 [24.5--44.5] & 15.37 [5.0--30.1] & 24.65 [17.3--31.8] \\
Task 2 average  & 12.86 [4.0--24.3] & 35.36 [27.2--43.7] & 14.65 [6.3--25.2] & 27.85 [21.1--34.5] \\
\bottomrule
\end{tabular}

\vspace{4pt}
\begin{minipage}{0.97\linewidth}
$^{*}$ At Center~3 the two constant references coincide, because that center's
training mode and the pooled training mode are both grade~3.
\end{minipage}
\end{table}

\begin{table}[htbp]
\centering
\tiny
\setlength{\tabcolsep}{4pt}
\renewcommand{\arraystretch}{1.05}
\caption{The twenty comparisons in which a method falls below a chance classifier on a given test set and metric. These cover eleven distinct method, test set and metric combinations, since a result may fall below both stochastic references, and the Task 2 average rows are not independent of the Centers 1 to 3 rows. The two parameter-efficient baselines are not included, as stated in Section~\ref{sec:baselines}. Values in percent; the chance level is the simulation mean from Table~\ref{tab:reference_classifiers}. For F1 $\uparrow$ a lower value is worse, for EC $\downarrow$ a higher value is worse.}
\label{tab:reference_comparisons}
\begin{tabular}{ll|l|rr}
\hline
\textbf{Test set} & \textbf{Method} & \textbf{Falls below} & \textbf{Method value} & \textbf{Chance level} \\
\hline
\multicolumn{5}{l}{\textit{Macro F1} $\uparrow$} \\
\hline
Task 2 Center 1 & Camma      & RandomUniform & 3.33 & 12.29 \\
Task 2 Center 1 & Cen        & RandomUniform & 11.43 & 12.29 \\
Task 2 Center 1 & SWARM      & RandomUniform & 10.32 & 12.29 \\
Task 2 Center 1 & Camma      & RandomPrior   & 3.33 & 14.78 \\
Task 2 Center 1 & Cen        & RandomPrior   & 11.43 & 14.78 \\
Task 2 Center 1 & SWARM      & RandomPrior   & 10.32 & 14.78 \\
Task 2 Center 2 & Elbflorenz & RandomUniform & 3.70 & 12.47 \\
Task 2 Center 2 & Elbflorenz & RandomPrior   & 3.70 & 13.80 \\
Task 2 Center 2 & Santhi     & RandomPrior   & 13.33 & 13.80 \\
Task 2 Center 3 & Santhi     & RandomUniform & 12.04 & 13.83 \\
Task 2 Center 3 & Santhi     & RandomPrior   & 12.04 & 15.37 \\
Task 2 average  & Elbflorenz & RandomUniform & 12.21 & 12.86 \\
Task 2 average  & Elbflorenz & RandomPrior   & 12.21 & 14.65 \\
Task 2 average  & Santhi     & RandomPrior   & 13.40 & 14.65 \\
Task 1 Center 4 & Camma      & RandomUniform & 4.76 & 11.91 \\
Task 1 Center 4 & Elbflorenz & RandomUniform & 7.83 & 11.91 \\
Task 1 Center 4 & Camma      & RandomPrior   & 4.76 & 14.30 \\
Task 1 Center 4 & Elbflorenz & RandomPrior   & 7.83 & 14.30 \\
\hline
\multicolumn{5}{l}{\textit{Expected cost} $\downarrow$} \\
\hline
Task 1 Center 4 & Camma      & RandomUniform & 57.24 & 33.55 \\
Task 1 Center 4 & Camma      & RandomPrior   & 57.24 & 25.52 \\
\hline
\end{tabular}
\end{table}

\section{Credit Authorship Contribution Statement}

\textbf{Max Kirchner:} Conceptualization, Challenge Organization, Methodology, Software, Validation, Formal Analysis, Investigation, Data Curation, Writing - Original Draft, Writing - Review and Editing, Visualization, Project Administration
\textbf{Hanna Hoffmann:} Conceptualization, Software, Writing - Review and Editing
\textbf{Alexander C. Jenke:} Challenge Organization, Software, Writing - Review and Editing
\textbf{Oliver L. Saldanha:} Conceptualization, Challenge Organization, Software, Investigation, Data Curation, Writing - Review and Editing
\textbf{Kevin Pfeiffer:} Software, Data Curation, Writing - Review and Editing
\textbf{Weam Kanjo:} Data Curation, Writing - Review and Editing
\textbf{Julia Alekseenko:} Methodology, Software, Investigation, Visualization, Writing - Review and Editing
\textbf{Claas de Boer:} Methodology, Software, Investigation, Writing - Review and Editing
\textbf{Santhi Raj Kolamuri:} Methodology, Software, Investigation, Writing - Review and Editing
\textbf{Lorenzo Mazza:} Methodology, Software, Investigation, Writing - Review and Editing
\textbf{Nicolas Padoy:} Challenge Team Supervision, Writing - Review and Editing
\textbf{Sophia Bano:} Challenge Organization, Supervision,  Writing - Review and Editing
\textbf{Annika Reinke:} Conceptualization, Challenge Organization, Supervision, Writing - Review and Editing
\textbf{Lena Maier-Hein:} Challenge Organization, Supervision,  Writing - Review and Editing
\textbf{Danail Stoyanov:} Challenge Organization, Supervision,  Writing - Review and Editing
\textbf{Jakob N. Kather:} Conceptualization, Challenge Organization, Writing - Review and Editing, Supervision, Project administration, Funding Acquisition
\textbf{Fiona R. Kolbinger:} Conceptualization, Challenge Organization, Investigation, Data Curation, Writing - Review and Editing, Supervision, Project Administration, Funding Acquisition
\textbf{Sebastian Bodenstedt:} Conceptualization, Challenge Organization, Software, Investigation, Data Curation, Writing - Review and Editing, Supervision, Project Administration, Funding Acquisition, Challenge Team Supervision
\textbf{Stefanie Speidel:} Conceptualization, Challenge Organization, Writing - Review and Editing, Supervision, Project Administration, Funding Acquisition, Challenge Team Supervision

\newpage

\bibliographystyle{elsarticle-num} 
\bibliography{references}



\end{document}